\documentclass[nohyperref]{article}

\usepackage{microtype}
\usepackage{graphicx}
\usepackage{subfigure}
\usepackage{booktabs} % for professional tables
\usepackage{enumitem}

\usepackage{hyperref}

\usepackage[accepted]{icml2022}

\usepackage{graphicx}
\usepackage{amsmath}
\usepackage{amssymb}
\usepackage{multirow}
\usepackage{subfigure}
\usepackage{enumitem}

\usepackage[capitalize,noabbrev]{cleveref}

\newcommand{\vect}[1]{\mathbf{#1}}

\newenvironment{tight_enumerate}{
	\begin{enumerate}[leftmargin=*]
		\setlength{\itemsep}{0pt}
		\setlength{\parskip}{0pt}
	}{\end{enumerate}}

\icmltitlerunning{Federated Unlearning: How to Efficiently Erase a Client in FL?}

\begin{document}

\twocolumn[
\icmltitle{Federated Unlearning: How to Efficiently Erase a Client in FL?}



\icmlsetsymbol{equal}{*}

\begin{icmlauthorlist}
\icmlauthor{Anisa Halimi}{ibme,equal}
\icmlauthor{Swanand Kadhe}{ibma,equal}
\icmlauthor{Ambrish Rawat}{ibme}
\icmlauthor{Nathalie Baracaldo}{ibma}
\end{icmlauthorlist}

\icmlaffiliation{ibme}{IBM Research Europe, Ireland}
\icmlaffiliation{ibma}{IBM Research, Almaden, USA}

\icmlcorrespondingauthor{Anisa Halimi}{anisa.halimi@ibm.com}
\icmlcorrespondingauthor{Swanand Kadhe}{swanand.kadhe@ibm.com}

\icmlkeywords{Machine Learning, ICML}

\vskip 0.3in
]



\printAffiliationsAndNotice{\icmlEqualContribution} % otherwise use the standard text.

\begin{abstract}
With privacy legislation empowering the users with the right to be forgotten, it has become essential to make a model amenable for forgetting some of its training data. However, existing unlearning methods in the machine learning context can not be directly applied in the context of distributed settings like federated learning due to the differences in learning protocol and the presence of multiple actors. In this paper, we tackle the problem of federated unlearning for the case of \emph{erasing a client} by removing the influence of their entire local data from the trained global model. To erase a client, we propose to first perform \emph{local unlearning} at the client to be erased, and then use the locally unlearned model as the initialization to run very few rounds of federated learning between the server and the remaining clients to obtain the unlearned \emph{global} model. We empirically evaluate our unlearning method by employing multiple performance measures on three datasets, and demonstrate that our unlearning method achieves comparable performance as the \emph{gold standard} unlearning method of federated retraining from scratch, while being significantly efficient. Unlike prior works, our unlearning method neither requires global access to the data used for training nor the history of the parameter updates to be stored by the server or any of the clients. 
\end{abstract}

\section{Introduction}\label{sec:introduction}
Modern machine learning is increasingly using large-size deep neural networks trained on massive datasets. While large model sizes and massive datasets have improved models performance, privacy implications have risen since large models tend to memorize aspects of their training data \cite{carlini2021extracting,lehman2021bert,carlini2023quantifying}. Such privacy implications make it challenging to meet privacy regulations \cite{voigt2017eu,pardau2018california,act2000personal}, which provide to data owners the right to be forgotten. Due to these privacy requirements, the field of \textit{machine unlearning} has recently received significant research attention \cite{nguyen2022survey,xu2023machine}. 

Machine unlearning, in a nutshell, removes the influence of specific training samples from a trained model, while maintaining the performance of the model. The problem of machine unlearning is even more challenging in the distributed paradigm of federated learning (FL), which allows multiple clients to jointly train a shared model while keeping their data on-premise~\cite{mcmahan2017communication}. Machine unlearning techniques developed for the centralized setting, e.g., ~\cite{graves2020amnesiac,bourtoule2021machine,guo2019certified} can not be directly applied in the FL setting due to its distributed nature, where all the participating clients contribute to learn the final \textit{global} model. A naive way of implementing federated unlearning is to retrain the model from scratch after removing from the corresponding client(s) the data sample(s) that are requested to be deleted. Retraining from scratch can be considered as the \textit{gold standard} for unlearning, as it can ensure \textit{exact} unlearning \cite{bourtoule2021machine,liu2021federaser}. However, it incurs prohibitively large communication and computation costs, making it infeasible in real-world FL settings.

In this paper, we tackle the problem of federated unlearning for the case when one of the clients wants to opt out of federation, and wants to remove the influence of their entire local data from the trained global model. We consider the so-called \textit{cross-silo} or \textit{enterprise} setting, wherein clients are different organizations (e.g., hospitals or banks) with strict privacy regulations \cite{kairouz2021fl}. We focus on erasing a single client because the cross-silo setup typically consists of a small number of clients and each client possesses significantly large amounts of data.

While there are a few works \cite{liu2021federaser,wu2022federated,wang2022federated,liu2022the} on federated unlearning, they either consider different setups than ours (e.g., unlearning an entire class or category) and/or require the server to store updates from each client in every round. In practical FL systems, client updates are only held ephemerally at the server because storing client updates at the server may have serious privacy implications due to potential leakage from model updates \cite{kairouz2021fl}. Therefore, our goal is to design a federated unlearning method that does not require the server or clients to store any client updates or even global updates. See Section \ref{sec:related_work} for a detailed comparison to prior works.

\textbf{Our Contributions.} 
\begin{tight_enumerate}
    \item We design an efficient federated unlearning method that erases a client by removing the influence of their entire local data from the trained global model. In our proposed method, the client to be erased first performs local unlearning by essentially \textit{reversing} the learning process. Next, by using the locally unlearned model as the initialization, the server and the remaining clients can obtain the global \textit{unlearned} model by performing very few rounds of federated learning.
    \item We empirically demonstrate that our unlearning method achieves comparable performance as the \textit{gold standard} of retraining from scratch, while being significantly efficient in terms of communication (and computation) costs. For instance, our method can reduce the communication cost compared to retraining by 5$\times$ to 24$\times$. We rigorously evaluate our unlearning method by employing three performance measures adapted from \citet{warnecke2023machine}: \textit{efficacy} (which measures success in removing the influence of data to be erased), \textit{fidelity} (which measures performance on data to be retained), and \textit{efficiency} (which measures costs compared to retraining from scratch). 
    \item Our key novelty is to formulate local unlearning problem as a constrained maximization problem, wherein the client to be erased maximizes their local loss while restricting the model parameters to an $\ell_2$-norm ball around a suitably chosen \textit{reference model} obtained from the other clients’ local models. Our formulation allows the client to efficiently perform local unlearning by using the Projected Gradient Descent (PGD). Starting with the \textit{locally} unlearned model enables the server and the remaining clients to obtain the \textit{global} unlearned model using very few FL rounds, resulting in significant efficiency gains over retraining.
\end{tight_enumerate}

\section{Related Work}\label{sec:related_work}
\noindent\textbf{Machine Unlearning.} The concept of machine unlearning, i.e., removing the impact of a data sample to the trained model, was first introduced by~\citet{cao2015towards}. After that, several algorithms for machine unlearning have been proposed~\cite{du2019lifelong,ginart2019making,guo2019certified,baumhauer2020machine,golatkar2020eternal,golatkar2020forgetting,graves2020amnesiac,bourtoule2021machine,neel2021descent,sekhari2021remember,thudi2021unrolling}. Such \textit{centralized} methods cannot be directly applied to FL due to its distributed nature, where no single participant has access to entire data.

\noindent\textbf{Federated Unlearning.} Unlearning in the FL setup has received relatively scant research attention, unlike the centralized setup. \citet{liu2021federaser, wu2022federated} consider a similar setup as ours, focusing on removing the contribution of a client after FL training. However, both these works require the server to store the updates from each client in every round. Storing model updates at the server may not be feasible in several application scenarios, especially with strict privacy regulations. In contrast, our method does not require the server or clients to store any client update or global update. In addition, \cite{wu2022federated} requires the server to possess some extra outsourced unlabeled data, which may not be realistic in several applications. Different from these works (and the one we propose), \citet{wang2022federated} propose an unlearning framework to forget a particular category or class. \citet{liu2022the} consider a setup where several clients want to erase small subset of their data, which is different from our setup. Further, their guarantees hold only for convex loss functions and their techniques require each client to compute an (approximate) inverse Hessian matrix which is computationally costly. In contrast, our unlearning techniques can be applied to non-convex objectives without substantial computational overhead.

\section{Background on Federated Learning}\label{sec:fl}
In a federated learning framework~\cite{mcmahan2017communication}, a (global) model is trained in a distributed way with the help of an aggregator (server), where each participating client contributes to training without sharing their data with the other participants. We consider the supervised federated learning setup with $N$ clients, each with dataset $\mathcal{D}_i = \{(\vect{x}_i, y_i)_{i\in[n_i]}\}$ (where $[n_i]=\{1,2,\dots,n_i\}$). The goal is to learn a model parameterized by weights $\vect{w}\in\mathbb{R}^d$. This is typically formulated as an empirical risk minimization problem: $\min_{\vect{w}\in\mathbb{R}^d}F(\vect{w}):=\sum_{i=1}^{N}p_i F_i(\vect{w})$, where $F_i(\cdot)$ is the local objective function at client $i$ and $p_i$ is the aggregation weight for client $i$. 

Federated learning systems typically use Federated Averaging (FedAvg)~\cite{mcmahan2017communication} for training. In round $t$, the server sends the current global model $\vect{w}^t$ to the clients. Each client takes multiple steps of mini-batch stochastic gradient descent (SGD) with a fixed learning rate to update their local model and sends their updated local model $\vect{w}^t_i$ and sends it to the server. Finally, the server computes a weighted average of the local models to obtain the global model for the next round: $\vect{w}^{t+1} = \sum_{i=1}^{N}p_i\vect{w}^t_i$, where $p_i = \frac{n_i}{\sum_{i=1}^{N}n_i}$. The iterative training process is repeated for a specific $T$ number of rounds.

We focus on the so-called \textit{enterprise} or \textit{cross-silo} setting in which clients are different organizations (e.g., banks or hospitals)~\cite{kairouz2021fl}. In this setting, the number of clients is often smaller, all the clients participate in each round, and every client possesses substantially large amount of data.

\section{Unlearning a Client}\label{sec:fun_challenges}

\subsection{Federated Unlearning Setup}\label{sec:setup}
We consider the following unlearning scenario in the FL setting. After FL training is performed with $N$ clients for the specified $T$ rounds, (Figure~\ref{fig:FUN-steps}(a)), a client $i\in[N]$ requests to opt out of federation and wants to remove the influence of their entire local data from the FL model. We refer to this client as the \textit{target client}. We focus on \textit{approximate unlearning} with the goal of obtaining a performance \textit{close to} retraining. 

Approximate unlearning relies on the fact that randomness in training induces a probability distribution over the models in the parameter space. At a high level, approximate unlearning ensures that the distribution of the unlearned model is either \textit{stochastically indistinguishable} from the distribution of the retrained model, where stochastic indistinguishability is typically characterized by using notions similar to differential privacy~\cite{guo2019certified,sekhari2021remember,warnecke2023machine}. It is possible to formalize theoretical notions of approximate federated unlearning, similar to those in the centralized setting.

While such theoretical notions allow for designing \textit{certified} unlearning algorithms, such algorithms are typically restricted to models with convex loss functions~\cite{guo2019certified,sekhari2021remember,warnecke2023machine}. On the other hand, practical FL systems often involve deep neural networks, which have non-convex loss~\cite{kairouz2021fl}. Therefore, we focus on the empirical evaluation of unlearning. In particular, we evaluate the unlearning algorithm by its \textit{efficacy, fidelity,} and \textit{efficiency} (see Section~\ref{sec:exp} for details). 

\subsection{Unlearning with Projected Gradient Descent}\label{sec:unlearning_method}
\begin{figure*}[ht!]
    \centering
    \includegraphics[scale=0.35]{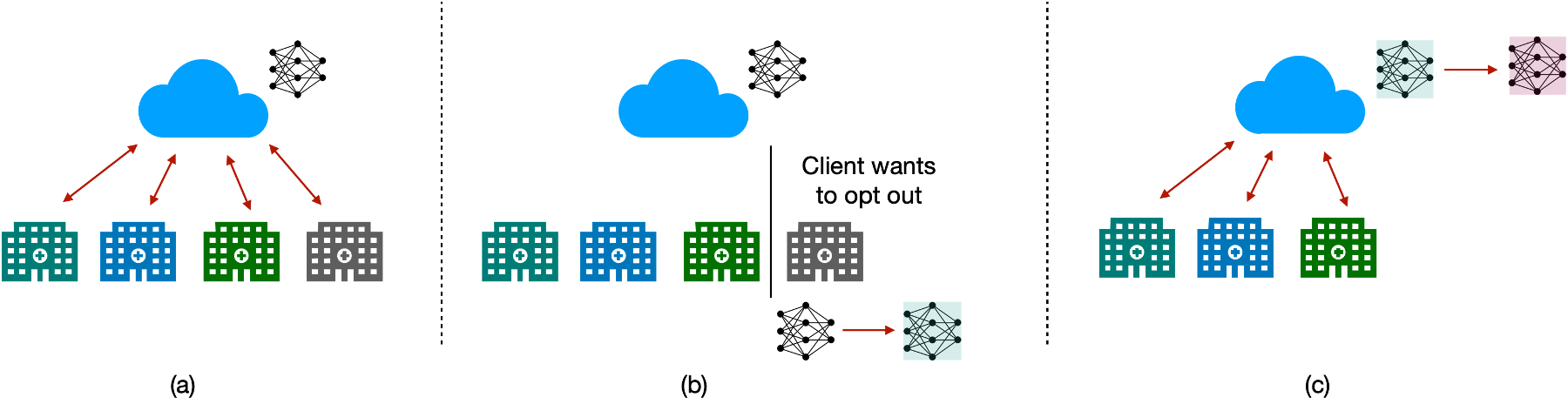}
    \caption{Phases of Federated Unlearning: (a) First, clients and the server participate in a federated learning process to train a global model. (b) One of the clients wants to opt out of the federation, and wants to unlearn their data. The target client $i$ locally runs Projected Gradient Descent (Algorithm~\ref{alg:PGA}) to obtain model $\vect{w}^u_i$. (c) The server and the remaining clients perform a few steps of federated learning with $\vect{w}^{u}_i$ as the initial point to obtain the final `unlearned' model (Algorithm~\ref{alg:PGA}).}
    \label{fig:FUN-steps}
    \vspace{-10pt}
\end{figure*}

As discussed in Section~\ref{sec:fl}, let $\vect{w}^T$ denote the global model after performing FL training for $T$ rounds. We propose to perform federated unlearning in two phases: (i) the target client $i$ performs local unlearning by essentially \textit{reversing} the learning process, (Figure~\ref{fig:FUN-steps}(b)), and (ii) the server and the retained clients start with the locally unlearned model, and perform a few rounds of federated learning to \textit{boost} its performance (Figure~\ref{fig:FUN-steps}(c)). We now describe in detail both these phases.

\noindent\textbf{Local Unlearning:} We argue that a natural idea for a client to unlearn their data is to \textit{reverse} this learning process. That is, during unlearning, instead of learning model parameters that minimize the empirical loss, the client strives to learn the model parameters to \textit{maximize} the loss. Indeed, prior works \cite{graves2020amnesiac,golatkar2020eternal,warnecke2023machine,jang2023knowledge} have applied \textit{gradient ascent} (or its variants) to find a model with \textit{large} empirical loss. However, these works restrict their attention to the case of unlearning only a handful (even just one) samples, whereas our focus is on unlearning the entire client dataset, which is typically large for the enterprise or cross-silo FL. In such cases, na\"ively applying gradient ascent to maximize the loss does not work because typical loss functions in practice are unbounded (e.g., cross-entropy loss). For an unbounded loss, each gradient ascent step moves towards a model that increases the loss, and after several steps, it is likely to produce an arbitrary model similar to a random model. Thus, we formulate unlearning at the target client as a constrained optimization problem and propose to solve it using \textit{projected gradient descent}.

To motivate our formulation, let us establish some notation for the federated training phase. In each round, the goal of a client is to learn a local model that \textit{minimizes} the (local) empirical risk, i.e., to solve the following problem: 
\begin{equation}
    \label{eq:train-ERM}
    \textrm{(Train)}\:\: \min_{\vect{w}\in\mathbb{R}^d}F_i(\vect{w}):=\frac{1}{n_i}\sum_{j\in \mathcal{D}_i}L(\vect{w};(\vect{x}_j,y_j)), 
\end{equation} 
where $L(\vect{w};(\vect{x}_j,y_j))$ is the loss of the prediction on example $(\vect{x}_j,y_j)$ made with model parameters $\vect{w}$. Each client locally makes several passes of (mini-batch stochastic) \textit{gradient descent} to find a model that has \textit{low} empirical loss. (It is also possible to use other optimization algorithms, e.g., Adam.)

During unlearning, we propose to ensure that the unlearned model is \textit{sufficiently close} to a \textit{reference model} that has effectively learned the other clients' data distributions. In particular, we propose to use the average of the other clients' models as a reference model, i.e., $\vect{w}_{\textrm{ref}} = \frac{1}{N-1}\sum_{j\ne i}\vect{w}^{T-1}_j$. Note that the target client $i$ can compute this reference model locally as $\vect{w}_{\textrm{ref}} = \frac{1}{N-1}\left(N\vect{w}^T - \vect{w}^{T-1}_i\right)$, where $\vect{w}^T$ is the global FL model after $T$ rounds and $\vect{w}^{T-1}_i$ is the $i$-th client's local model update in round $T-1$. The client $i$ then optimizes over the model parameters that lie in the $\ell_2$-norm ball of radius $\delta$ around $\vect{w}_{\textrm{ref}}$. (The radius $\delta$ will be treated as a hyperparameter in our experiments.) Thus, during unlearning, the client solves the following optimization problem:
\begin{equation}
    \label{eq:unlearning-ERM}
    \textrm{(Unlearn)}\:\: \max_{\vect{w}\in\{\vect{v}\in\mathbb{R}^d:\lVert\vect{v}-\vect{w}_{\textrm{ref}} \rVert_2 \leq \delta\}}F_i(\vect{w}),
\end{equation} 
where $F_i(\cdot)$ is defined in eq.~\eqref{eq:train-ERM}.

A natural choice for solving~\eqref{eq:unlearning-ERM} is to use \textit{projected gradient descent}. More specifically, let us denote the $\ell_2$-norm ball of radius $\delta$ around $\vect{w}_{\textrm{ref}}$ as $\Omega = \{\vect{v}\in\mathbb{R}^d:\lVert\vect{v}-\vect{w}_{\textrm{ref}} \rVert_2 \leq \delta\}$. Let $\mathcal{P}:\mathbb{R}^d\rightarrow\mathbb{R}^d$ denote the projection operator onto $\Omega$. Then, for a given step-size $\eta_u$, client $i$ uses projected gradient descent (PGD)\footnote{Note that eqn.~\eqref{eq:PGD-update} is technically projected gradient \textit{ascent} since we are maximizing a function rather than minimizing. However, similar to adversarial machine learning literature (see, e.g.,~\citet{madry2018towards}), we refer to the process as projected gradient descent.} to iterate the update:
\begin{equation}
    \label{eq:PGD-update}
    \mathbf{w} \leftarrow \mathcal{P}\left(\vect{w} + \eta_u\nabla F_i(\vect{w};b)\right),
\end{equation}
where $\nabla F_i(\vect{w;b})$ is the gradient of $F_i$ with respect to $\vect{w}$ computed on a batch $b$.
To avoid learning an arbitrary model, we perform early stopping if the $\ell_2$-distance of the target client $\vect{w}^{T-1}_i$ to the unlearned model $\vect{w}^{u}_i$ is smaller than a predetermined threshold $\tau$ (which is treated as a hyperparameter). Algorithm~\ref{alg:PGA} describes the local unlearning procedure, and Appendix 1 provides a schematic.

\noindent\textbf{FL post-training.} To improve the performance of the locally unlearned model on the data of the retained clients, the server and the retained clients perform a few rounds of FL training starting with the unlearned model $\vect{w}^u_i$. The detailed steps are described in Algorithm~\ref{alg:PGA}. Interestingly, we demonstrate empirically in Section~\ref{sec:exp} that performing \textit{very few rounds} of FL post-training on the unlearned model $\vect{w}^u_i$ gives good performance in practice.

%%%%%%%%%%%%%%%%%%%%%%%%%%%%%%%%%%%%%%%%%%%%%%%%%%%%%%%
% Unlearning Algo
%%%%%%%%%%%%%%%%%%%%%%%%%%%%%%%%%%%%%%%%%%%%%%%%%%%%%%
\begin{algorithm}[!t]
\caption{Federated Unlearning}
\label{alg:PGA}
\begin{algorithmic}
\STATE {\bfseries Local Unlearning at Client} $i$ {\bfseries via Projected Gradient Descent}: 
\STATE Inputs: learning rate $\eta_u$, batch size $B_u$, number of epochs $E_u$, clipping radius $\delta$, and early stopping threshold $\tau$
\STATE Set $\vect{w}_{\textrm{ref}} \leftarrow \frac{1}{N-1}\left(N\vect{w}^T - \vect{w}^{T-1}_i\right) = \frac{1}{N-1}\sum_{i\ne j}\vect{w}^{T-1}_j$
\STATE Define $\mathcal{P}(\vect{w})$ as the projection of $\vect{w}\in\mathbb{R}^d$ onto the $\ell_2$-norm ball $\Omega = \{\vect{v}\in\mathbb{R}^d : \lVert \vect{v} - \vect{w}_{\textrm{ref}} \rVert\leq \delta\}$
\STATE Initialize unlearning model as $\vect{w} \leftarrow \vect{w}_{\textrm{ref}}$
\STATE $\mathcal{B}_i \leftarrow$ (split $\mathcal{D}_i$ into batches of size $B_u$)
%(split $D_i^{tr}$ into batches of size $B_u$)
\FOR{each local epoch $e=1$ {\bfseries to} $E_u$}
    \FOR{batch $b$ in $\mathcal{B}_i$}
        \STATE $\vect{w} \leftarrow \mathcal{P}\left(\vect{w} + \eta_u \nabla F_i(\vect{w};b)\right)$  
        %\IF{$\textrm{Accuracy}(\vect{w};D^{Val}) < \tau$}
        \IF{$\lVert\vect{w} - \vect{w}^{T-1}_i\rVert_2 < \tau$}
            \STATE Set $\vect{w}^{u}_i \leftarrow \vect{w}$
            and return $\vect{w}^{u}_i$ to server
        \ENDIF
    \ENDFOR
\ENDFOR
\STATE Set $\vect{w}^{u}_i \leftarrow \vect{w}$
            and return $\vect{w}^{u}_i$ to server
\STATE {}
\STATE {\bfseries FL post-training:}
\STATE Inputs: learning rate $\eta_p$, batch size $B_p$, number of epochs $E_p$, number of rounds $T_p$
\STATE {\bfseries Server executes}
\STATE Initialize $\vect{w}^0 \leftarrow \vect{w}^u_i$
\FOR{each round $t = 1$ {\bfseries to} $T_p$}
\STATE Send $\vect{w}^{t-1}$ to clients $[N]\setminus \{i\}$
    \FOR{each client $j\in [N]\setminus \{i\}$ {\bfseries in parallel}}
        \STATE $\vect{w}^t_j \leftarrow$ ClientUpdate$(j, \vect{w}^{t-1})$
    \ENDFOR
    \STATE $\vect{w}^t \leftarrow \sum_{j}\frac{n_j}{\sum_{l}n_l}\vect{w}^t_j$
\ENDFOR
\STATE Set the unlearned model as $\vect{w}^u \leftarrow \vect{w}^{T_p}$
\STATE {}
\STATE {\bfseries ClientUpdate}$(j, \vect{w}^{t-1})$:
\STATE $\mathcal{B}_j \leftarrow$ (split $\mathcal{D}_j$ into batches of size $B_p$)
\STATE Initialize $\vect{w} \leftarrow \vect{w}^{t-1}$
\FOR{each local epoch $e=1$ {\bfseries to} $E_p$}
    \FOR{batch $b$ in $\mathcal{B}_j$}
        \STATE $\vect{w} \leftarrow \vect{w} - \eta_p \nabla F_j(\vect{w};b)$ 
    \ENDFOR
\ENDFOR
\STATE Return $\vect{w}$ to the server
\end{algorithmic}
\end{algorithm}
%%%%%%%%%%%%%%%%%%%%%%%%%%%%%%%%%%%%%%%%%%%%%%%%%%%%%%

\section{Evaluation}\label{sec:exp}

\noindent\textbf{Unlearning Scenarios:} We consider two scenarios to illustrate the phenomenon of unlearning: (i) removing the effect of backdoor triggers and (ii) removing the effect of flipping. At a high level, a successful federated unlearning method should produce a model that does not perform well on the target client's data distribution while keeping good performance on the other clients' data distribution. The goal in the above mentioned scenarios is to deliberately differentiate the target client's data distribution from the data distribution of the other clients. 

\noindent\textbf{Performance Measures:}
In general, an effective federated unlearning method must remove the contribution of the target client's data, maintain good performance, and be more efficient than retraining from scratch. To reflect these properties in our evaluation, we use three performance measures (similar to \citet{warnecke2023machine}).

\noindent\textit{Efficacy of unlearning.} 
The efficacy of an unlearning method measures how successful it is in removing the contribution of the target client's data. We quantify the efficacy of unlearning by evaluating the performance of the unlearned model on the target client's data distribution. In particular, we use the following two metrics to measure efficacy: (i) accuracy on the target client's data distribution: depending on the scenario this will be accuracy on a hold-out test set of backdoored or flipped images; and (ii) membership inference risk with respect to the target client's dataset. 

\noindent\textit{Fidelity of unlearning.} 
The fidelity of an unlearning method measures whether it can maintain a performance close to the original model. We quantify the fidelity of unlearning by evaluating the performance of the unlearned model on the retained clients' data distribution. In particular, we use the accuracy of the unlearned model on a hold-out test set of clean images to measure fidelity. 

\noindent\textit{Efficiency of unlearning.} While it is straightforward to perform unlearning by retraining the FL model from scratch without the participation of the target client, such retraining incurs significant communication and computation costs. The efficiency of an unlearning method measures the reduction in communication and computation costs wrt. retraining. We evaluate the efficiency of the proposed unlearning method by comparing its communication cost with that of retraining. We focus on the communication cost since it is known to be a key bottleneck in FL~\cite{kairouz2021fl}.

\noindent\textbf{Datasets and Model Architecture:} To evaluate the performance of the proposed method, we utilize three datasets: MNIST~\cite{lecun1998mnist}, EMNIST (balanced version)~\cite{cohen2017emnist}, and CIFAR-10~\cite{krizhevsky2009learning}. For all datasets, we use a CNN from~\cite{mcmahan2017communication} with two $5\times 5$ convolution layers, a fully connected layer with 512 units and ReLu activation, and a final softmax output layer ($1,663,370$ total parameters). We equally partition the training images of each dataset across $N$ clients in the FL process, one of which is the target client. We detail the hyperparameters used for FL training and unlearning in Appendix~\ref{app:hyperparameters}.

\subsection{Unlearning Scenario 1: Backdoors}\label{sec:backdoored_images}

We use the backdoor triggers~\cite{gu2017badnets} as an effective way to evaluate the performance of unlearning methods, similar to~\citet{wu2022federated}. In particular, the target client uses a dataset in which a certain fraction of images has a backdoor trigger inserted in them. Because of this client, the global FL model becomes susceptible to the backdoor trigger. Then, a successful unlearning process should produce a model that reduces the accuracy of the images with the backdoor trigger, while maintaining high accuracy on regular (clean) images. For backdoors, we introduce a `pixel pattern' trigger of size $3\times 3$ using the Adversarial Robustness Toolbox~\cite{art2018}, and change the label of corresponding samples to `9' for MNIST, to `t' for EMNIST, and to `truck' for CIFAR-10. When inserting the backdoor trigger, we exclude the data sample whose label is already the target label. We consider two cases: (i) $N=5$ clients with the target client having $66\%$ of their images backdoored, and (ii) $N=10$ clients with the target client having $80\%$ of their images backdoored. We compare our proposed unlearning method to retraining from scratch. 

\begin{figure*}[ht!]
    \centering
    \begin{subfigure}[MNIST]{\includegraphics[scale=0.28]{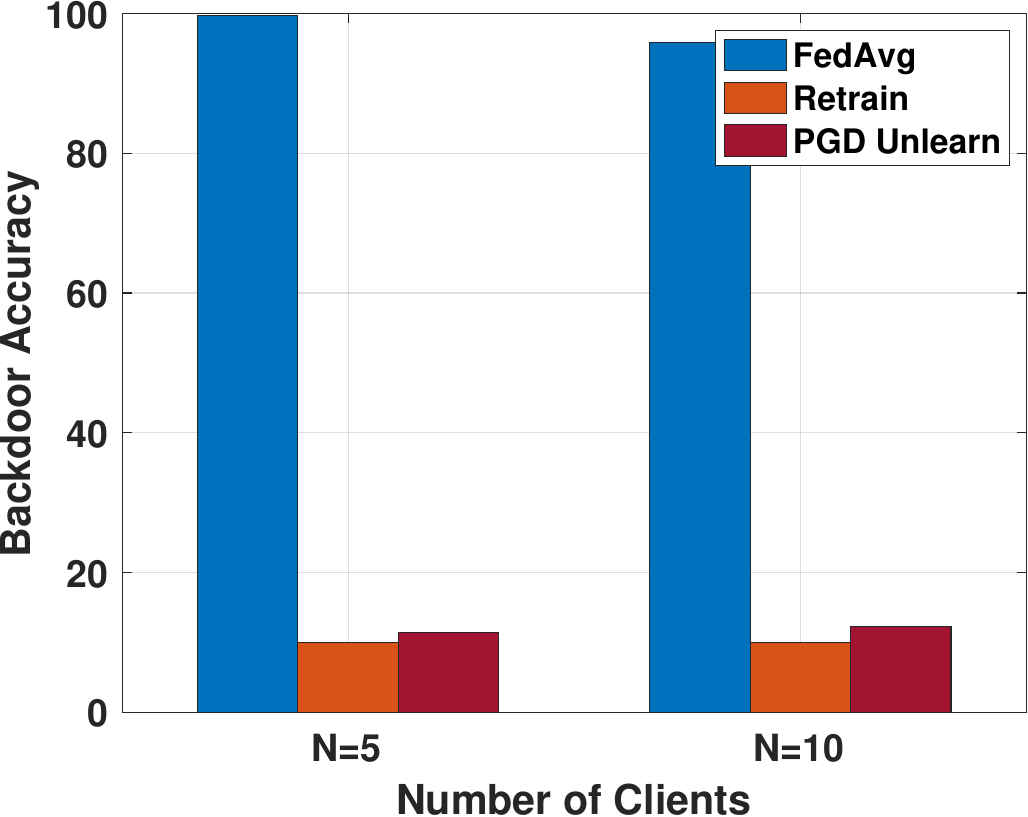}}
    \end{subfigure}\hfill
    \begin{subfigure}[EMNIST]{\includegraphics[scale=0.28]{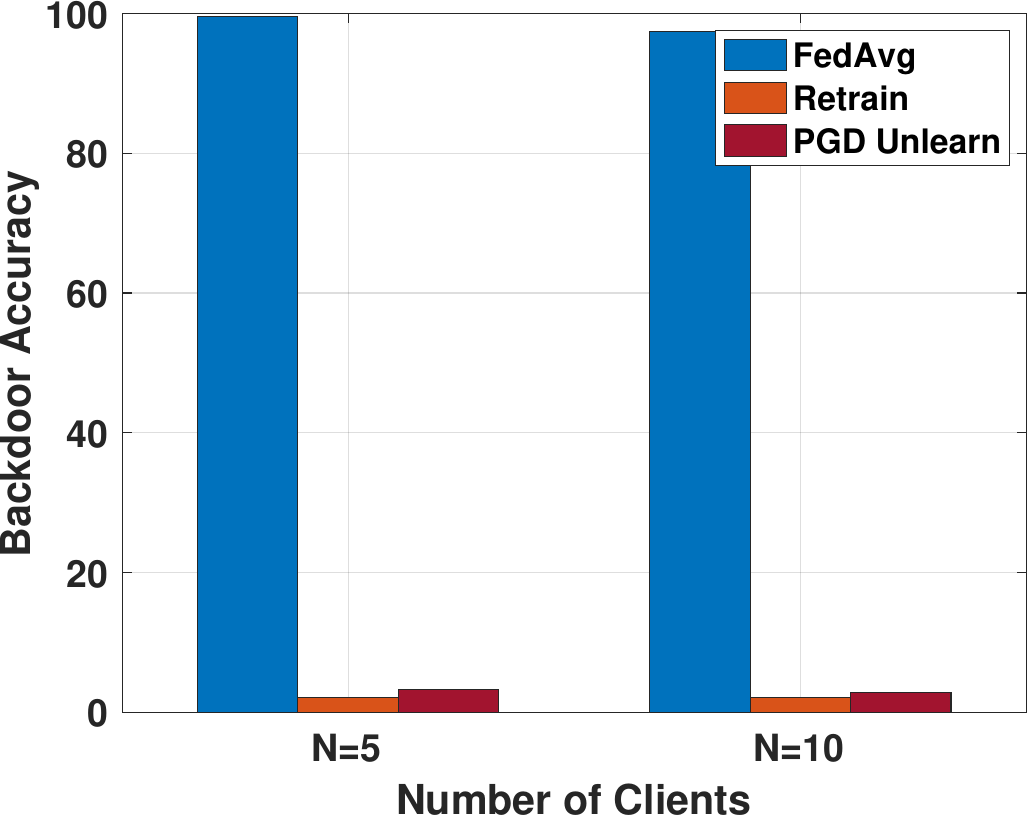}}
    \end{subfigure}\hfill
    \begin{subfigure}[CIFAR-10]{\includegraphics[scale=0.28]{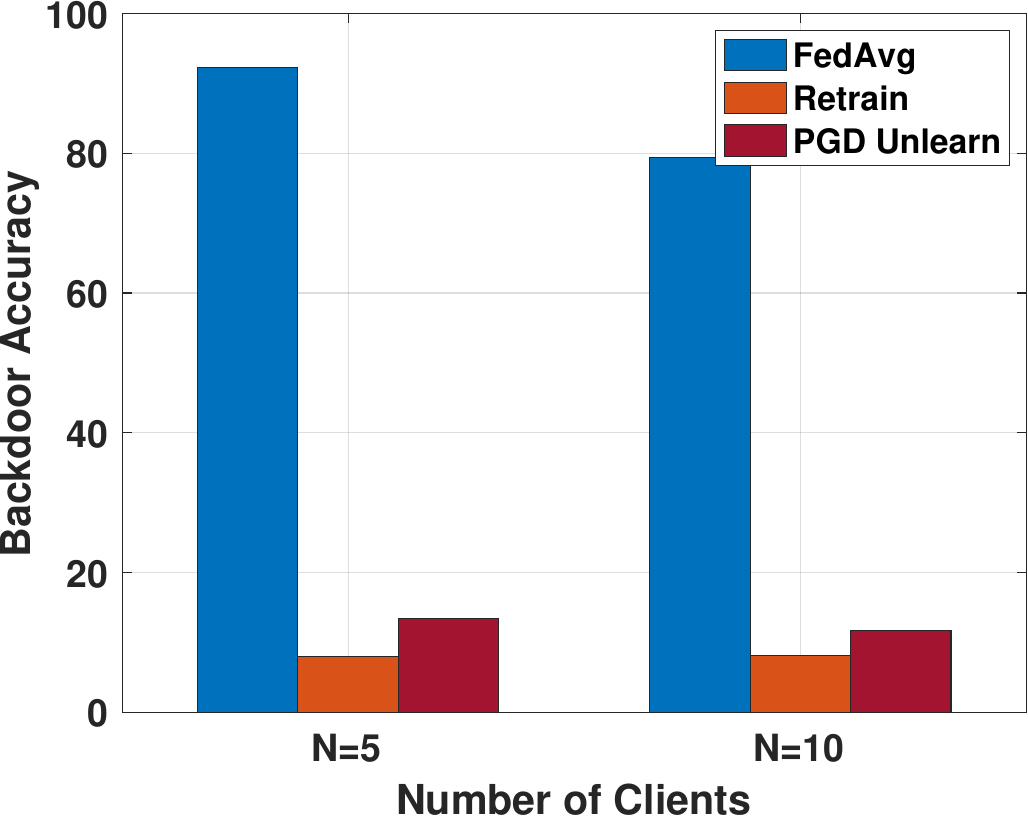}}
    \end{subfigure}\hfill
    \vspace{-5pt}
    \caption{Backdoor accuracy (efficacy) of the fully retrained and the PGD-based unlearned model in each dataset, and their comparison with the FedAvg model before unlearning. The backdoor accuracy of the PGD-based unlearned model is obtained after $1$ round of FL post-training. Our method significantly reduces the backdoor accuracy compared to FedAvg model and achieves a similar performance as retraining, which demonstrates its high unlearning efficacy. 
    }
    \label{fig:backdoor_acc}
    \vspace{-5pt}
\end{figure*}

\noindent\textbf{Efficacy evaluation:} We analyze the efficacy of the unlearning method using two metrics. First, we evaluate the accuracy on a hold-out test set of backdoored images. We compute the accuracy on the backdoored data (referred as the backdoor accuracy) as the percentage of triggered data that are misclassified as the target label required by the attacker. The lower the backdoor accuracy, the better the model has unlearned the contribution of the target client's data. Figure~\ref{fig:backdoor_acc} shows the backdoor accuracy of each model for each dataset for both cases. For both cases and all datasets, the high value of backdoor accuracy for the FedAvg model indicates that the FL model has learned the target client's data consisting of backdoor triggers. We observe that our proposed PGD-based unlearning method substantially reduces the backdoor accuracy, and in fact, achieves similar backdoor accuracy to retraining for all datasets. This demonstrates that the efficacy of our method in terms of backdoor accuracy is comparable to that of retraining. 

\begin{figure*}[ht!]
    \centering
    \begin{subfigure}[MNIST]{\includegraphics[scale=0.37]{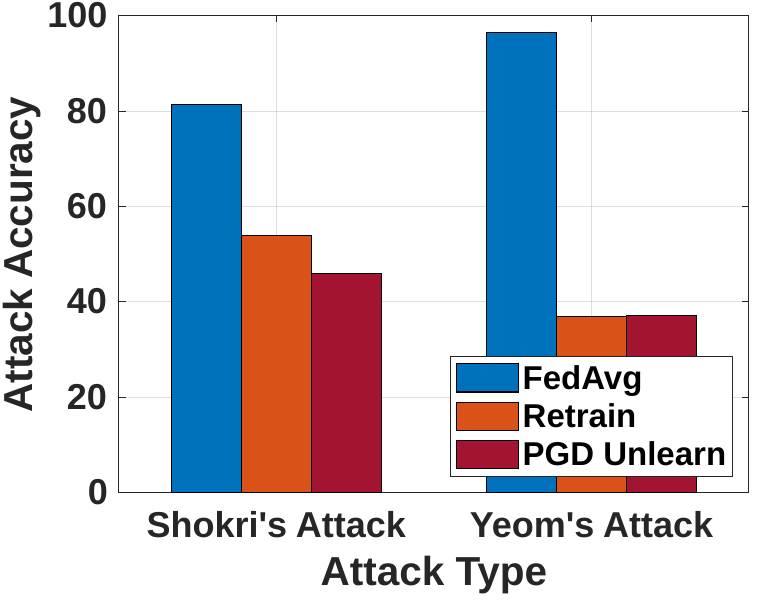}}
    \end{subfigure}\hfill
    \begin{subfigure}[EMNIST]{\includegraphics[scale=0.37]{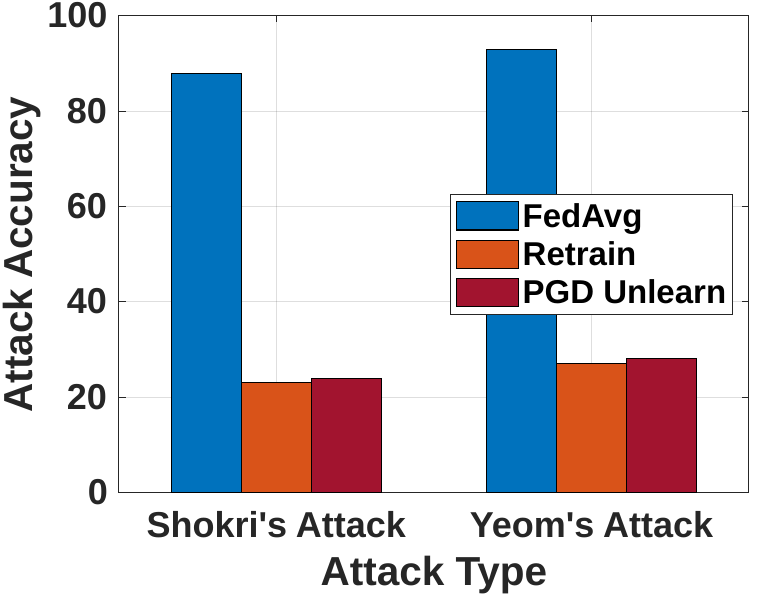}}
    \end{subfigure}\hfill
    \begin{subfigure}[CIFAR-10]{\includegraphics[scale=0.37]{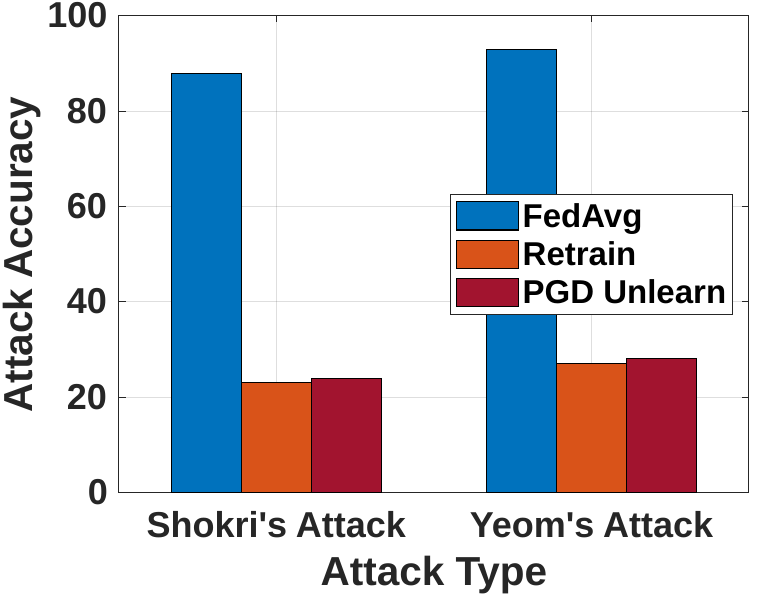}}
    \end{subfigure}\hfill
    \vspace{-5pt}
    \caption{Backdoor Scenario: Membership inference attacks accuracy (efficacy) for the two attacks and the three datasets for $N=5$ clients. Our proposed method achieves a similar attack accuracy as retraining, which demonstrates its high efficacy.}
    \label{fig:mem_inf_num_parties_5}
    \vspace{-5pt}
\end{figure*}
Another metric that we use to measure the efficacy of the proposed unlearning method is the membership inference risk. The goal of a membership inference attack is to determine whether a specific data sample is part of the dataset used to train the model. We leverage membership inference attacks to assess how much information from the target client's data is part of the unlearned model, similar to~\citet{liu2021federaser}. A successful unlearning process should produce a model that has a low membership inference risk on the data of the target client. To measure the membership inference risk, we use two well-known membership inference attacks: \citet{shokri2017membership}, which uses the idea of training \textit{shadow models}; and \citet{yeom2018privacy}, which uses training and test time loss values. A description of the attacks is provided in Appendix~\ref{app:mem_inf}.

For the evaluation, we perform the attacks against the PGD-based unlearned model (after one round of post-training), the fully retrained model, and the FedAvg model (the global model obtained by FL before any unlearning). For simplicity, in Shokri's attack, we use the FedAvg model as the shadow model. We compute the attack accuracy as the percentage of the target client's data that are inferred as being part of the training dataset. Figure~\ref{fig:mem_inf_num_parties_5} shows the accuracy of the membership inference attacks for $N=5$ clients. We observe that both unlearning methods achieve substantially lower membership inference attack accuracies than the one in the FedAvg model for all datasets and both attacks. In fact, for both attacks, the proposed unlearning approach obtains similar accuracy to retraining, which demonstrates the high efficacy of our method. 

\begin{figure*}[ht!]
    \centering
    \begin{subfigure}[MNIST]{\includegraphics[scale=0.28]{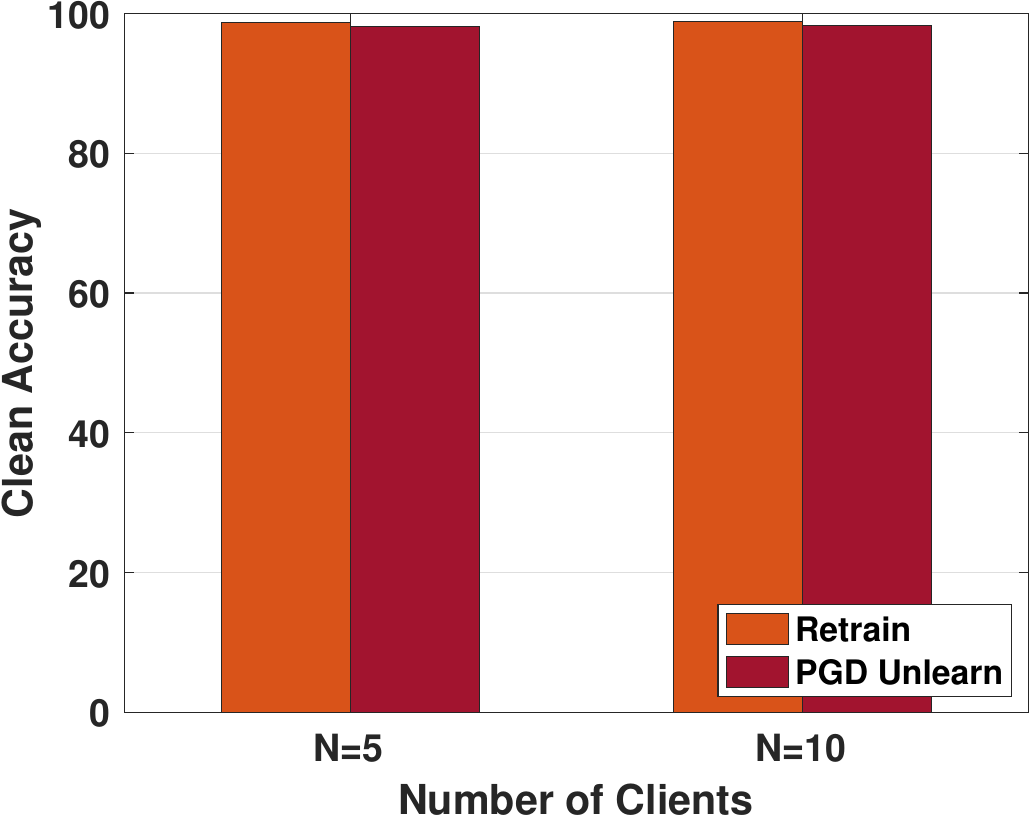}}
    \end{subfigure}\hfill
    \begin{subfigure}[EMNIST]{\includegraphics[scale=0.28]{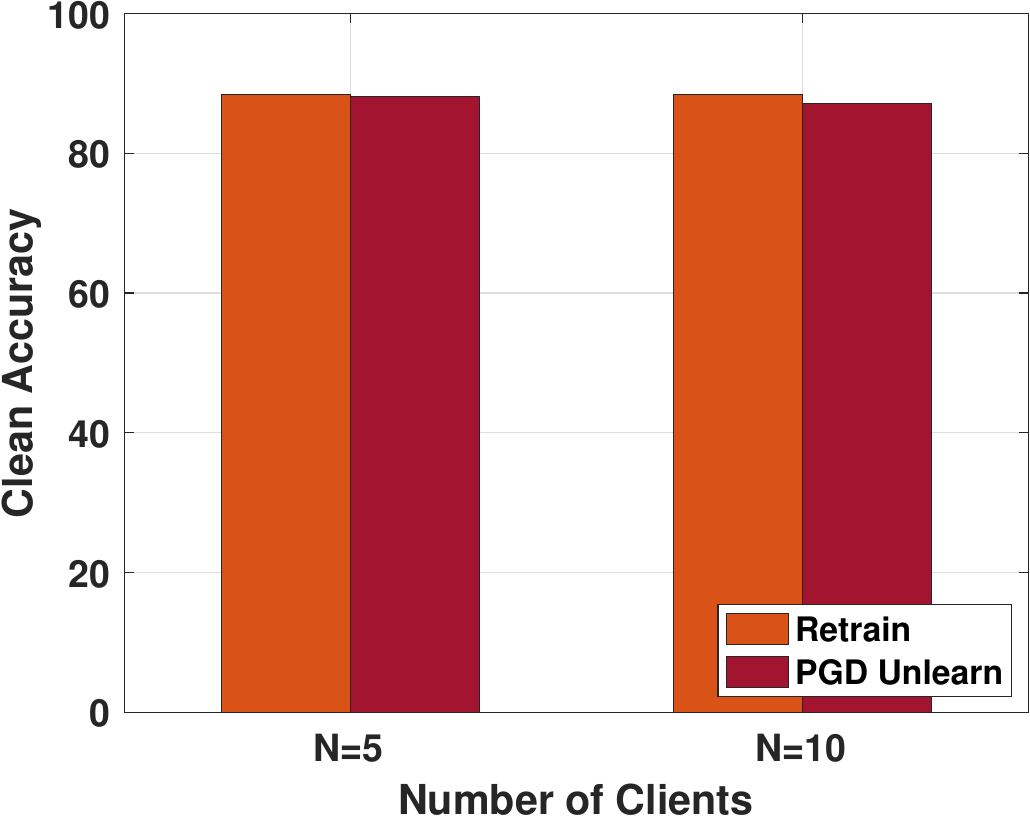}}
    \end{subfigure}\hfill
    \begin{subfigure}[CIFAR-10]{\includegraphics[scale=0.28]{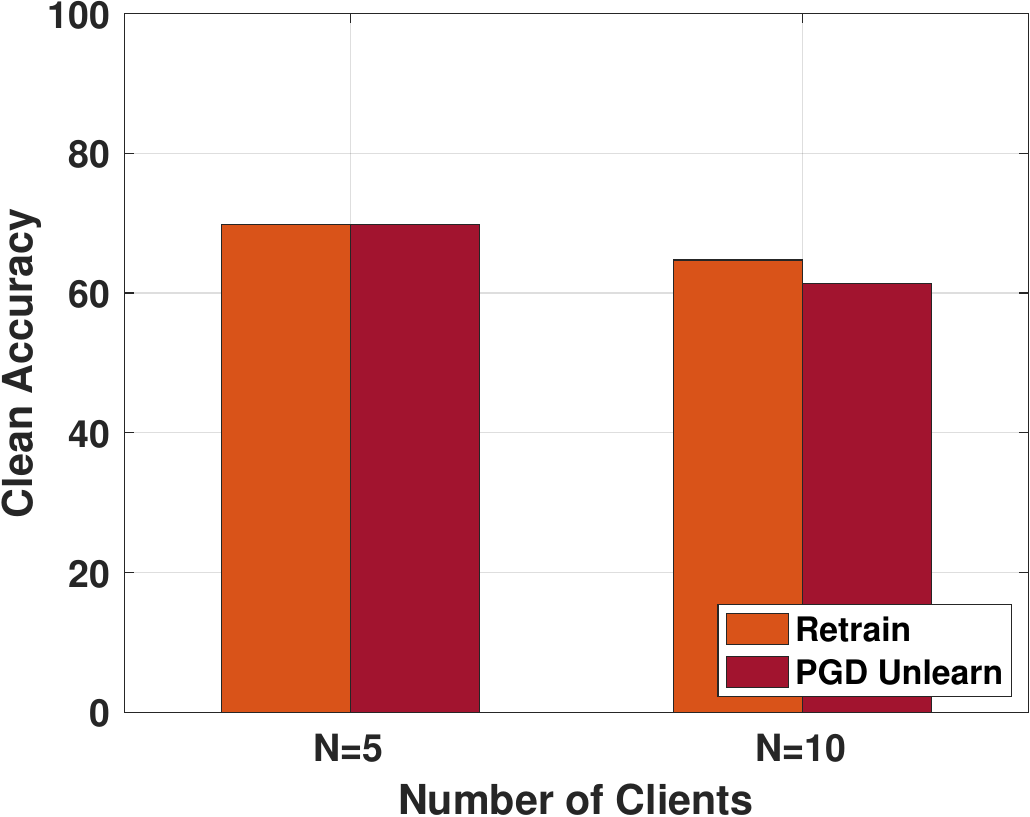}}
    \end{subfigure}\hfill
    \vspace{-5pt}
    \caption{Backdoor Scenario: Clean accuracy of the fully retrained and the PGD-based unlearned model in each dataset. The clean accuracy of the PGD-based unlearned model is obtained after $5$ rounds of FL post-training. Our unlearning method achieves similar clean accuracy to retraining, which demonstrates its high fidelity.}
    \label{fig:clean_acc}
    \vspace{-5pt}
\end{figure*}

\noindent\textbf{Fidelity evaluation:} We evaluate the fidelity of our unlearning method by computing the accuracy of the unlearned model on a hold-out test set that consists of clean images (no backdoor triggers). We refer to the accuracy computed on the clean images as clean accuracy. Note that the clean images represent the data distribution of the retained clients, and the clean accuracy indicates whether the unlearned model can maintain good performance on the retained data. In Figure~\ref{fig:clean_acc}, we show the clean accuracy of the unlearned models obtained by our method and retraining. We observe that our PGD-based unlearning method achieves similar clean accuracy to retraining, demonstrating the capability of our method for unlearning with high fidelity. 

\begin{figure*}[ht!]
    \centering
    \begin{subfigure}[MNIST]{\includegraphics[scale=0.28]{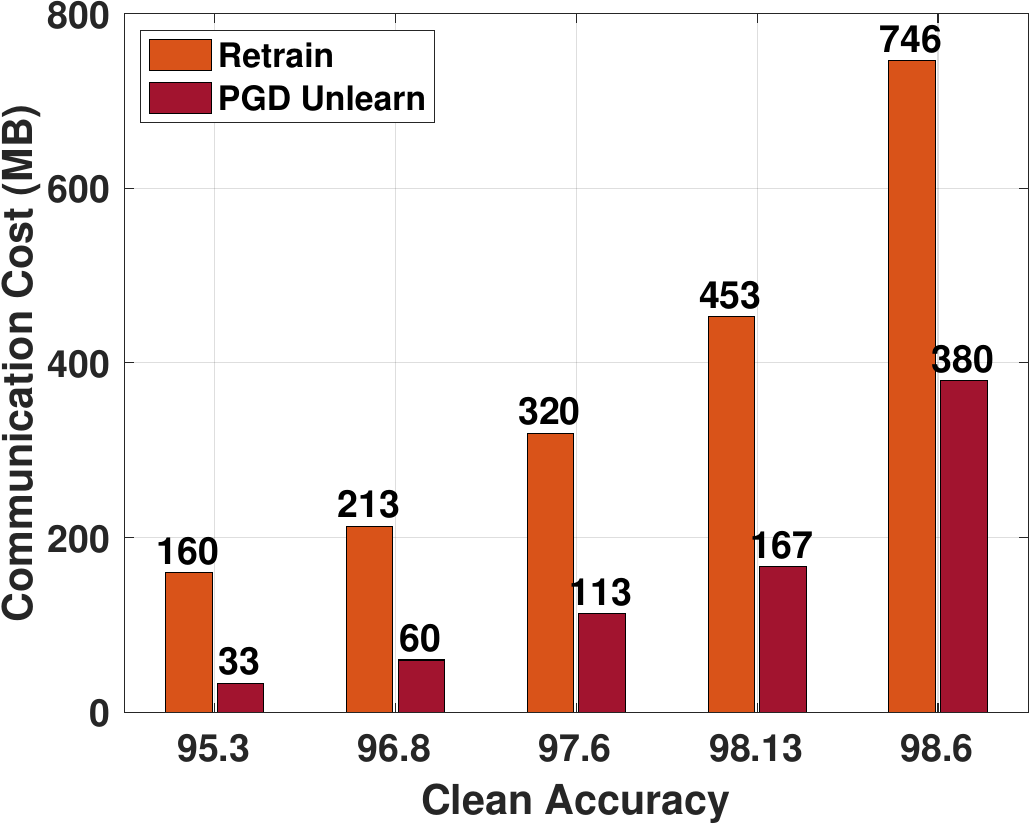}}
    \end{subfigure}\hfill
    \begin{subfigure}[EMNIST]{\includegraphics[scale=0.28]{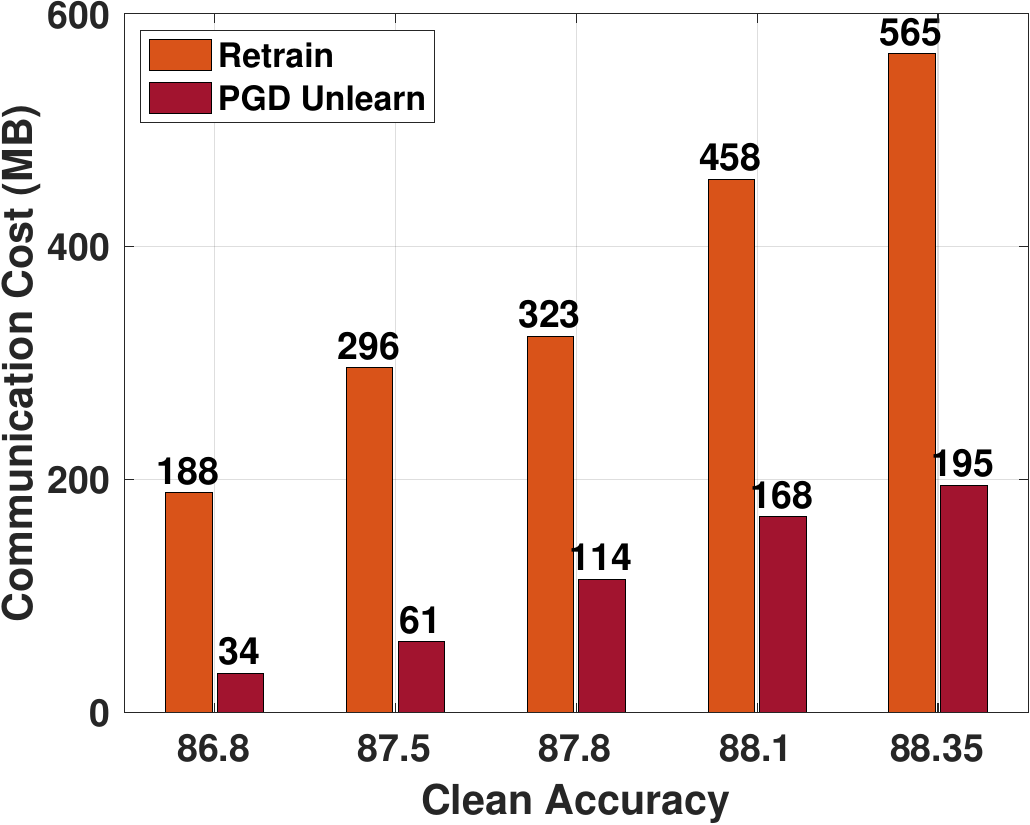}}
    \end{subfigure}\hfill
    \begin{subfigure}[CIFAR-10]{\includegraphics[scale=0.28]{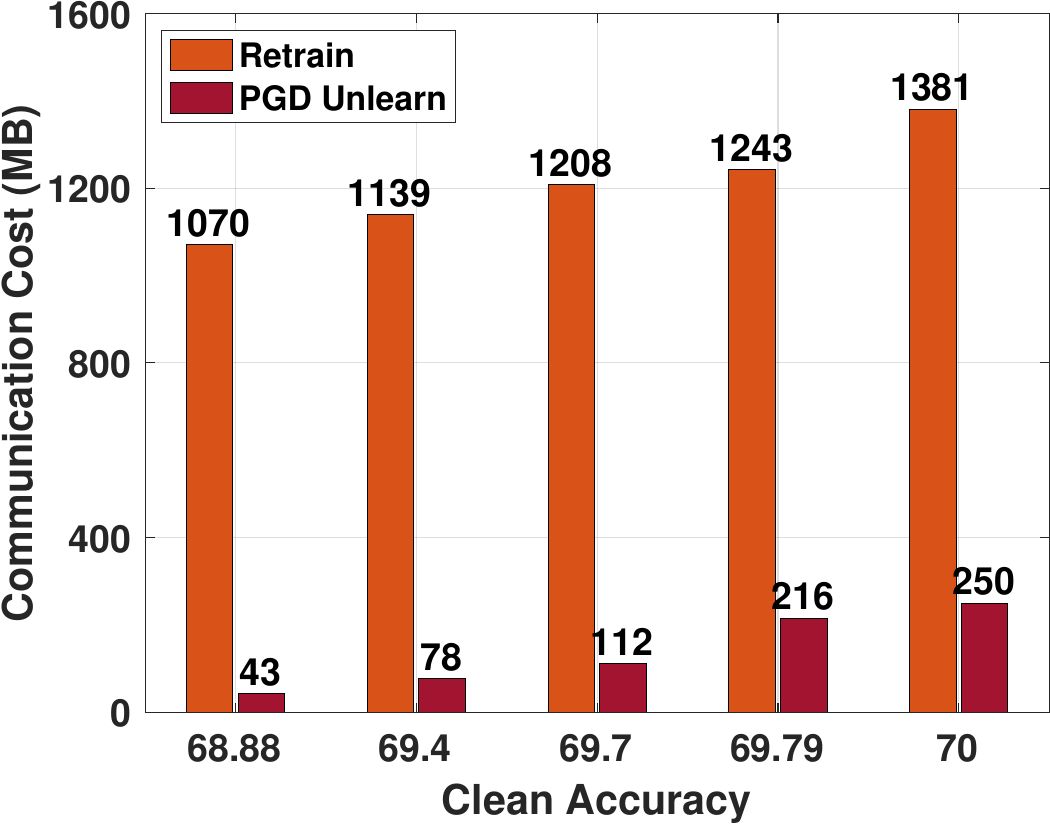}}
    \end{subfigure}\hfill
    \vspace{-5pt}
    \caption{Backdoor Scenario: Communication costs (efficiency) of the proposed unlearning method and the baseline approach of retraining with respect to the clean accuracy (fidelity) in each dataset for $N=5$ clients.}
    \label{fig:unlearn_comm_costs_num_parties_5}
    \vspace{-5pt}
\end{figure*}
\begin{figure*}[ht!]
    \centering
    \begin{subfigure}[MNIST]{\includegraphics[scale=0.28]{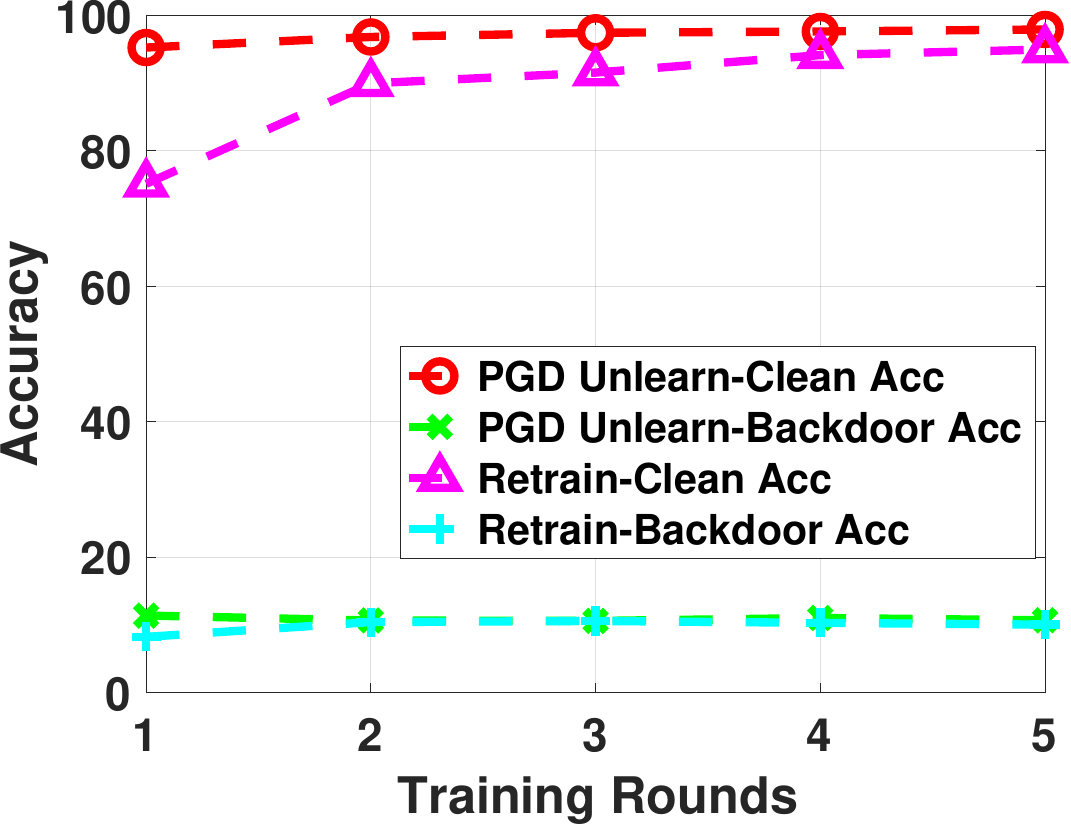}}
    \end{subfigure}\hfill
    \begin{subfigure}[EMNIST]{\includegraphics[scale=0.28]{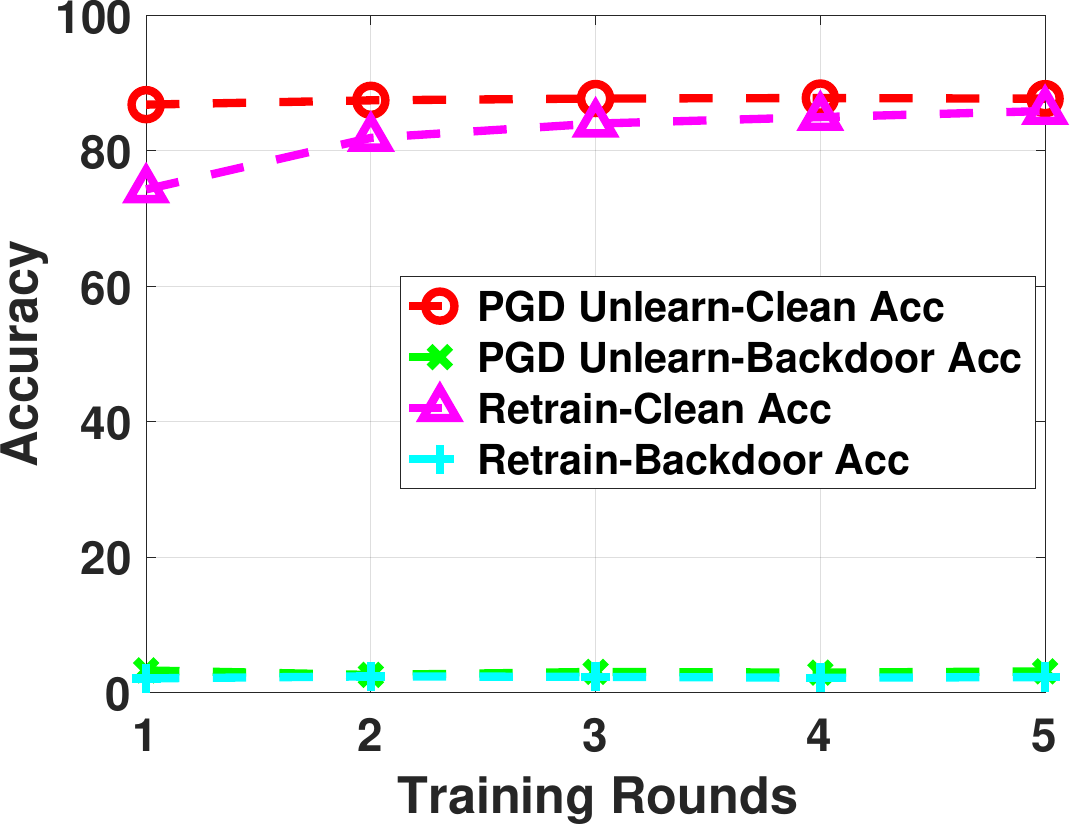}}
    \end{subfigure}\hfill
    \begin{subfigure}[CIFAR-10]{\includegraphics[scale=0.28]{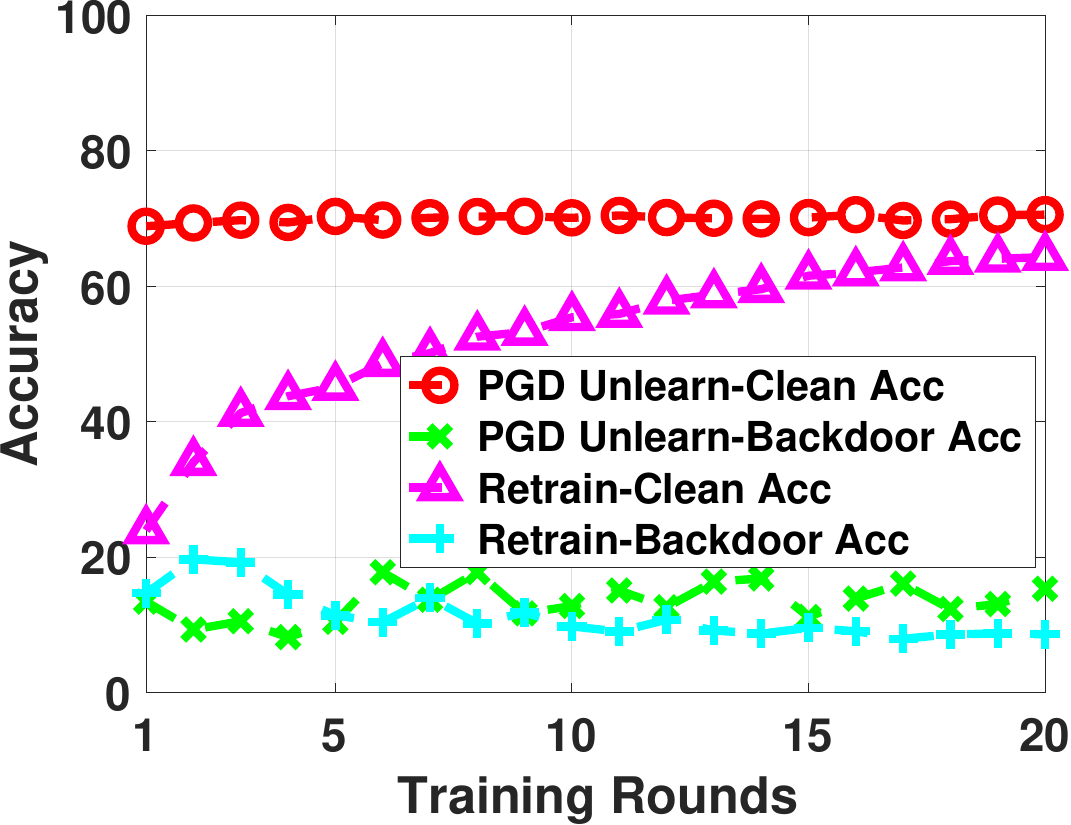}}
    \end{subfigure}\hfill
    \vspace{-5pt}
    \caption{Clean accuracy (fidelity) and backdoor accuracy (efficacy) of the PGD-based unlearned and fully retrained model with respect to the number of rounds in each dataset for $N=5$ clients.}
    \label{fig:unlearn_client_rounds_num_parties_5}
    \vspace{-5pt}
\end{figure*}
\noindent\textbf{Efficiency evaluation:} To evaluate the efficiency of our method we compare its communication cost with retraining. We compute the communication cost of a given approach as the total size of the model updates (in MB) that clients participating in the FL process communicate to the server. Figure~\ref{fig:unlearn_comm_costs_num_parties_5} shows the communication cost for various clean accuracy (fidelity) values for $N=5$ clients. We observe that the proposed unlearning method is significantly more efficient than retraining while achieving similar fidelity. For instance, in the MNIST dataset, to reach a fidelity (clean accuracy) of $98.13\%$, our method requires $167$ MB of communication cost, whereas retraining from scratch requires $453$ MB. Thus, our method is $2.7 \times$ more efficient in terms of communication costs than retraining. This gap is even higher for the EMNIST and CIFAR-10 datasets. Overall, we observe that our proposed unlearning method reduces communication costs by up to $24 \times$. 

To compare the speedup of the proposed method to the baseline of retraining, we compute the clean and backdoor accuracy of both methods with respect to the number of FL training rounds. Note that, for our method, the FL training starts with the locally unlearned model that the target client has obtained using the projected gradient descent (as discussed in Section~\ref{sec:unlearning_method}). On the other hand, for the baseline of retraining, the FL training starts with a randomly initialized model. Figure~\ref{fig:unlearn_client_rounds_num_parties_5} shows this comparison for $N=5$ clients. We provide the results for $N=10$ clients in Appendix~\ref{app:backdoor_acc_number_rounds}. 
After one round of post-training, the proposed method reaches a clean accuracy of $95.3\%$ and a backdoor accuracy of $11.38\%$ in the MNIST dataset. Retraining requires more than $5$ training rounds to achieve similar performance. This shows that the PGD-based local unlearning produces an effective starting point by removing the influence of the target client's data without degrading the performance on the other clients' data. 

\subsection{Unlearning Scenario 2: Flipped Images}\label{sec:flipped_images}
In this scenario, we consider a target client that has a dataset where a certain fraction of images are flipped. We do not apply any data augmentation on any clients' datasets. A successful unlearning method should reduce the accuracy on the flipped images while maintaining high accuracy on regular images. For evaluation, we consider two cases: (i) $N=5$ clients with the target client having $66\%$ of their images flipped and (ii) $N=10$ clients with the target client having $80\%$ of their images flipped. We flip the images horizontally and keep the label unchanged. Note that for this unlearning scenario, we do not provide the results for the CIFAR-10 dataset for the following reason. The accuracy of retraining on a hold-out test set of flipped images for $N=5$ is $71.14\%$. This value is similar to the accuracy on a hold-out test set of regular images (no flipping applied), making CIFAR-10 an inappropriate dataset for this scenario. 

\begin{figure}[ht!]
    \centering
    \begin{subfigure}[MNIST]{\includegraphics[scale=0.21]{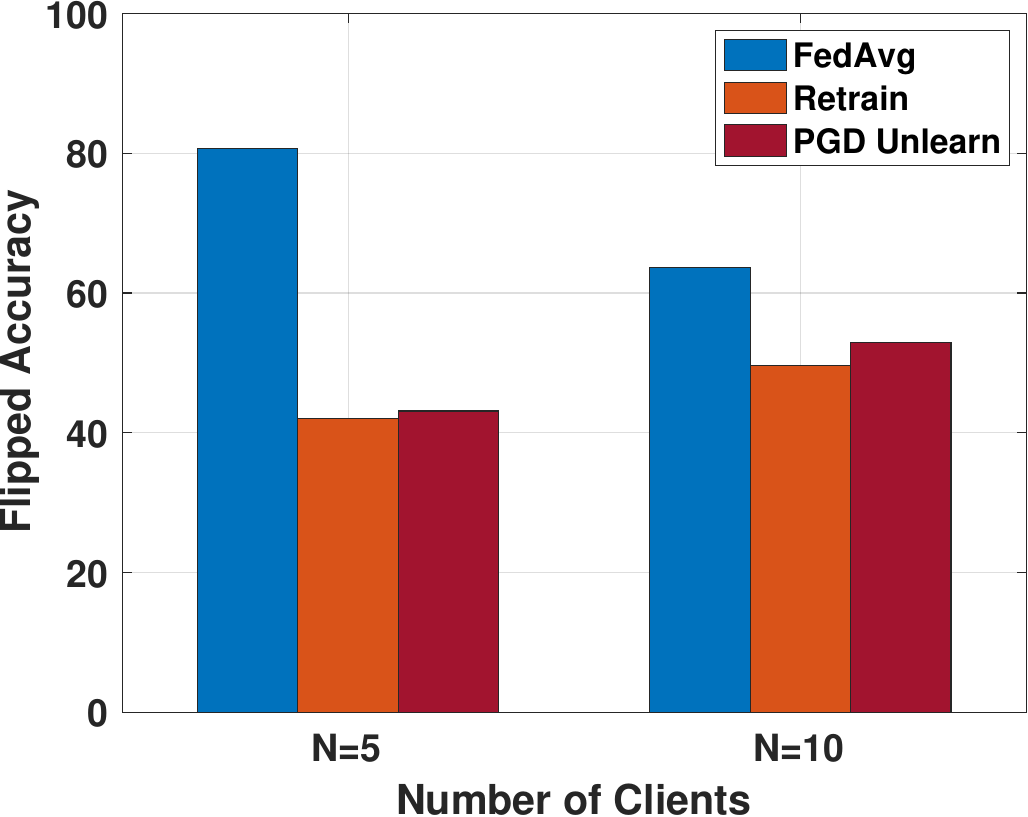}}
    \end{subfigure}\hfill
    \begin{subfigure}[EMNIST]{\includegraphics[scale=0.21]{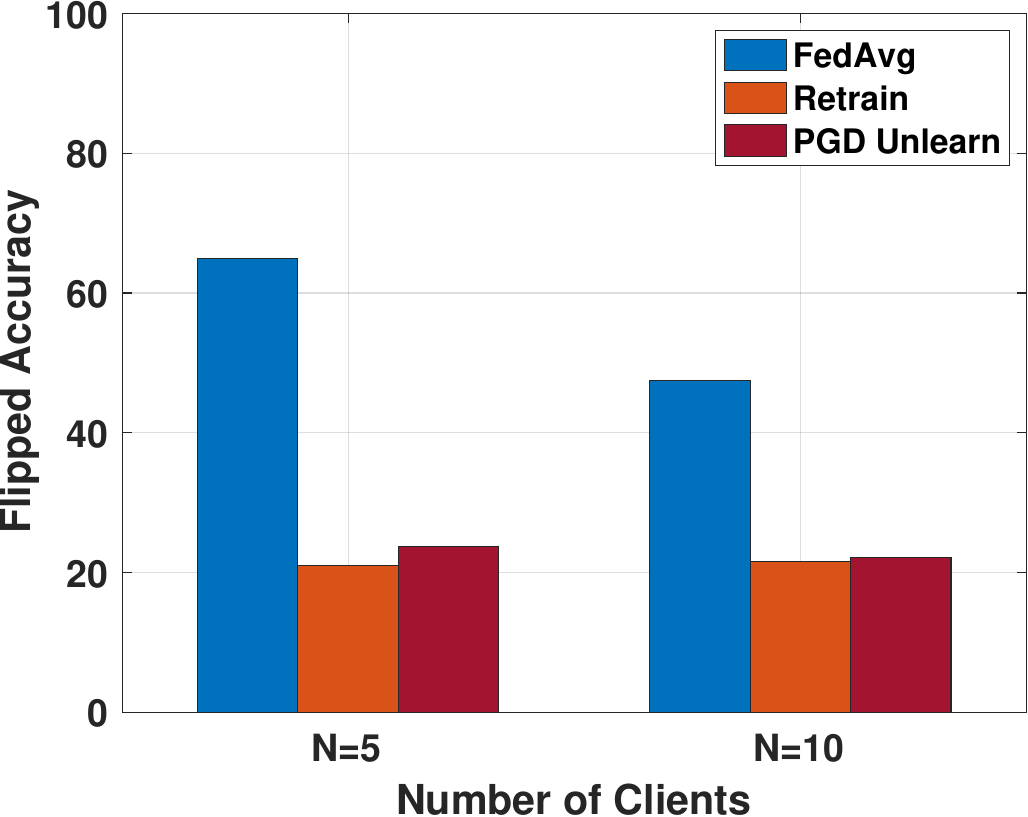}}
    \end{subfigure}\hfill
    \vspace{-5pt}
    \caption{Flipped accuracy of the FedAvg (before unlearning), fully retrained, and PGD-based unlearned models in the MNIST and EMNIST datasets. The flipped accuracy of the PGD-based unlearned model is obtained after $5$ rounds of FL post-training. Our method substantially reduces the flipped accuracy compared to the FedAvg model and achieves similar performance as retraining, which demonstrates its high unlearning efficacy.}
    \label{fig:flipped_acc}
    \vspace{-5pt}
\end{figure}

\noindent\textbf{Efficacy evaluation:} To analyze the efficacy of the unlearning methods, we use the accuracy on a hold-out test set of flipped images (referred as the flipped accuracy). In Figure~\ref{fig:flipped_acc}, we show the flipped accuracy of the FedAvg, fully retrained, and the PGD-based unlearned models in the MNIST and EMNIST datasets for both cases. We observe that the proposed unlearning method achieves a similar flipped accuracy to retraining. These results show that our PGD-based unlearning method has comparable efficacy to retraining for removing the contribution of the target client's data.

\begin{figure}[ht!]
    \centering
    \begin{subfigure}[MNIST]{\includegraphics[scale=0.21]{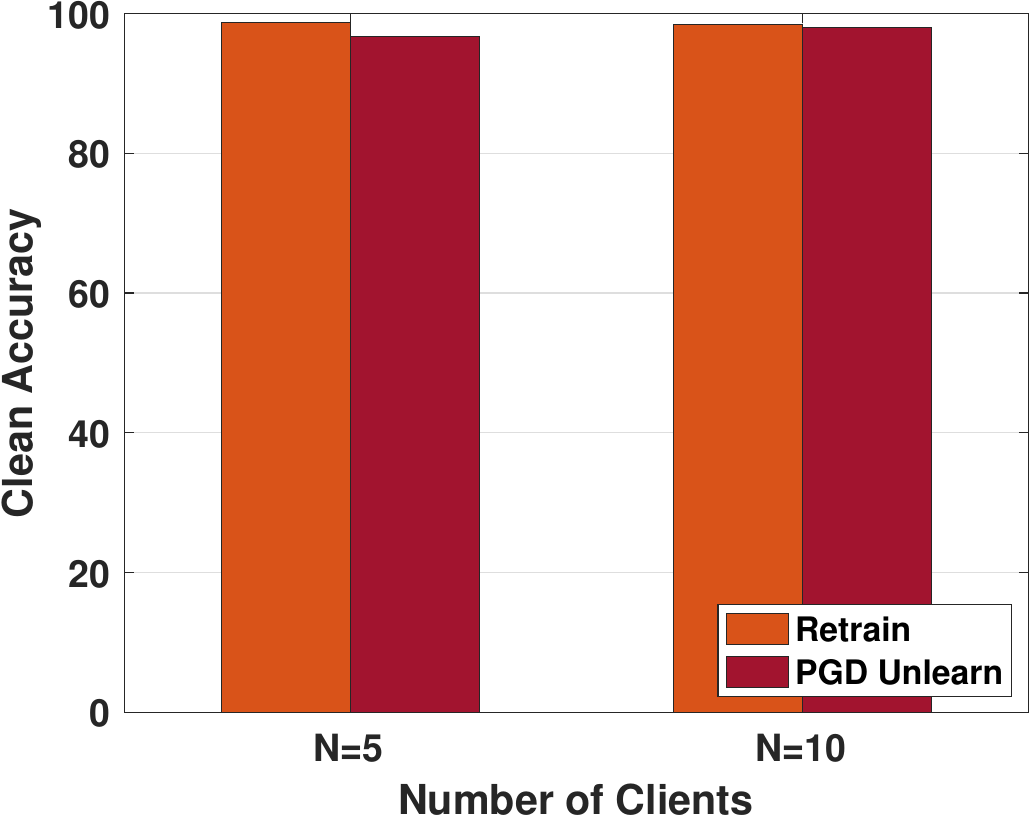}}
    \end{subfigure}\hfill
    \begin{subfigure}[EMNIST]{\includegraphics[scale=0.21]{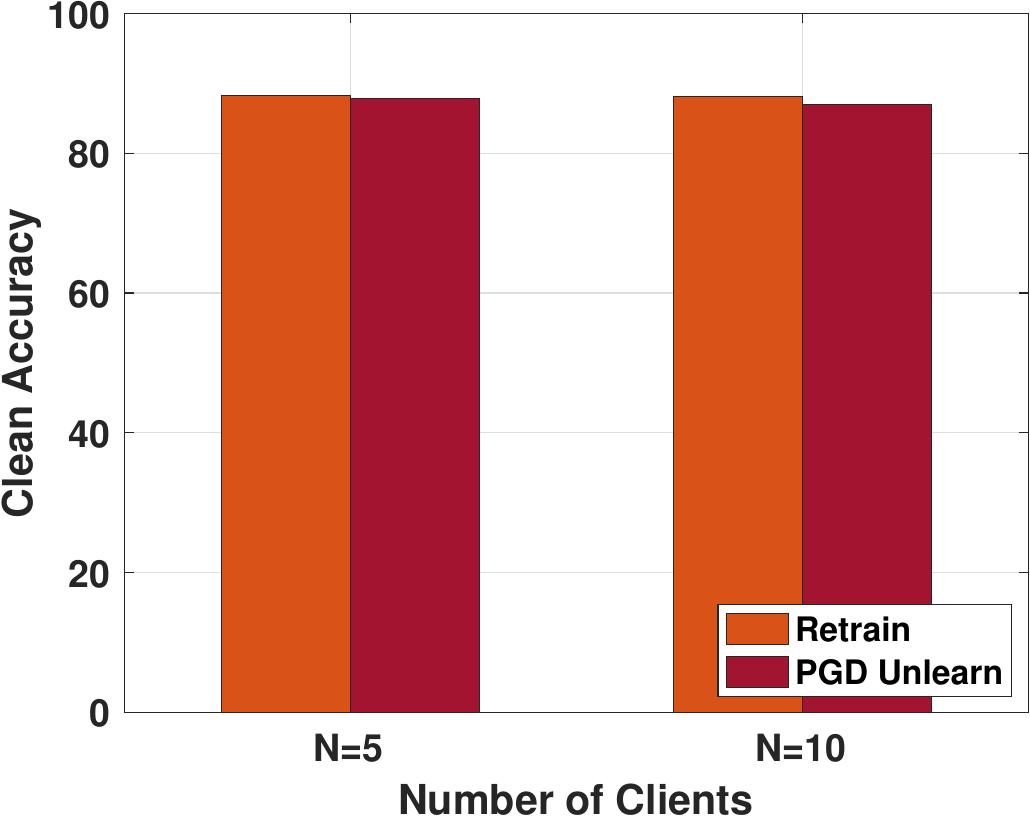}}
    \end{subfigure}\hfill
    \vspace{-5pt}
    \caption{Flipping Scenario: Clean accuracy (fidelity) of the fully retrained and the PGD-based unlearned models in the MNIST and EMNIST datasets. The clean accuracy of the PGD-based unlearned model is obtained after $5$ rounds of FL post-training.}
    \label{fig:flip_clean_acc}
    \vspace{-5pt}
\end{figure}

\noindent\textbf{Fidelity evaluation:} To evaluate the fidelity of the unlearning methods, we examine the accuracy between the proposed unlearning method and the baseline approach on a hold-out test set of regular images (no flipping applied). For consistency, we refer to the accuracy on regular images as the clean accuracy. Figure~\ref{fig:flip_clean_acc} shows the clean accuracy of the fully retrained and PGD-based unlearned models in each dataset for both cases. We observe that our method maintains a high clean accuracy, which is similar to the baseline approach of retraining. Since the clean images represent the data distribution of the retained clients, the high clean accuracy of our method indicates that it can perform unlearning with high fidelity by maintaining good performance on the retained data.

\begin{figure}[ht!]
    \centering
    \begin{subfigure}[MNIST]{\includegraphics[scale=0.21]{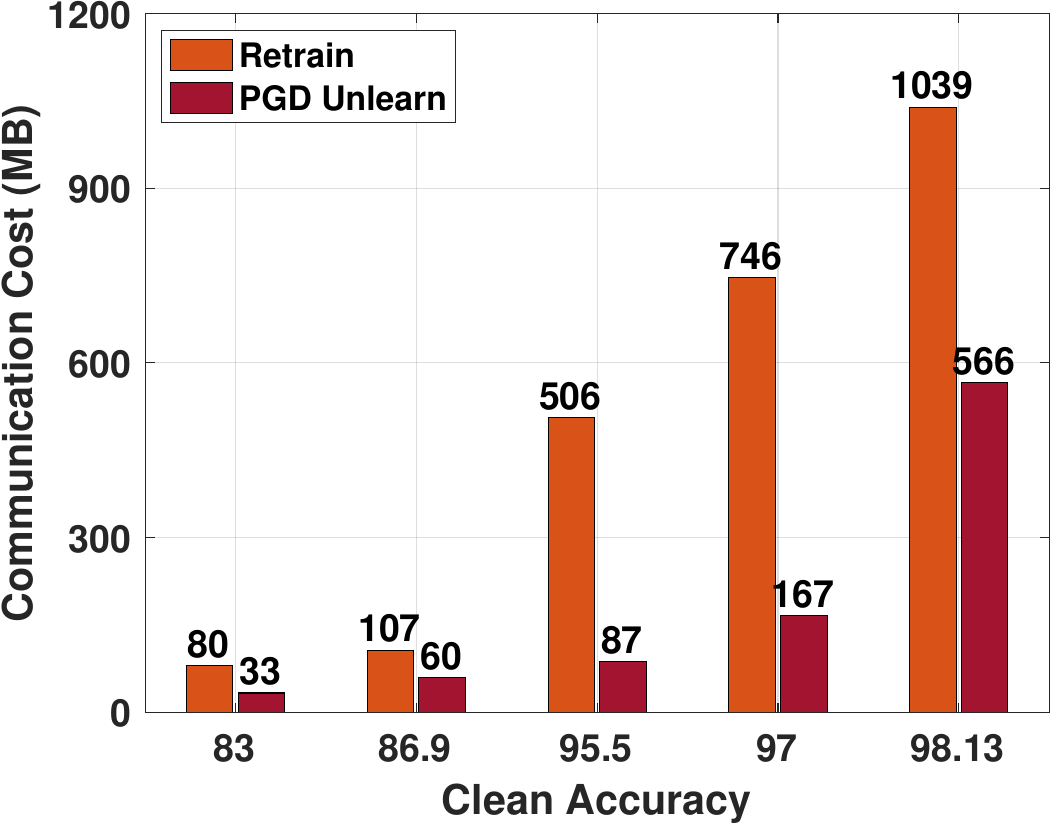}}
    \end{subfigure}\hfill
    \begin{subfigure}[EMNIST]{\includegraphics[scale=0.21]{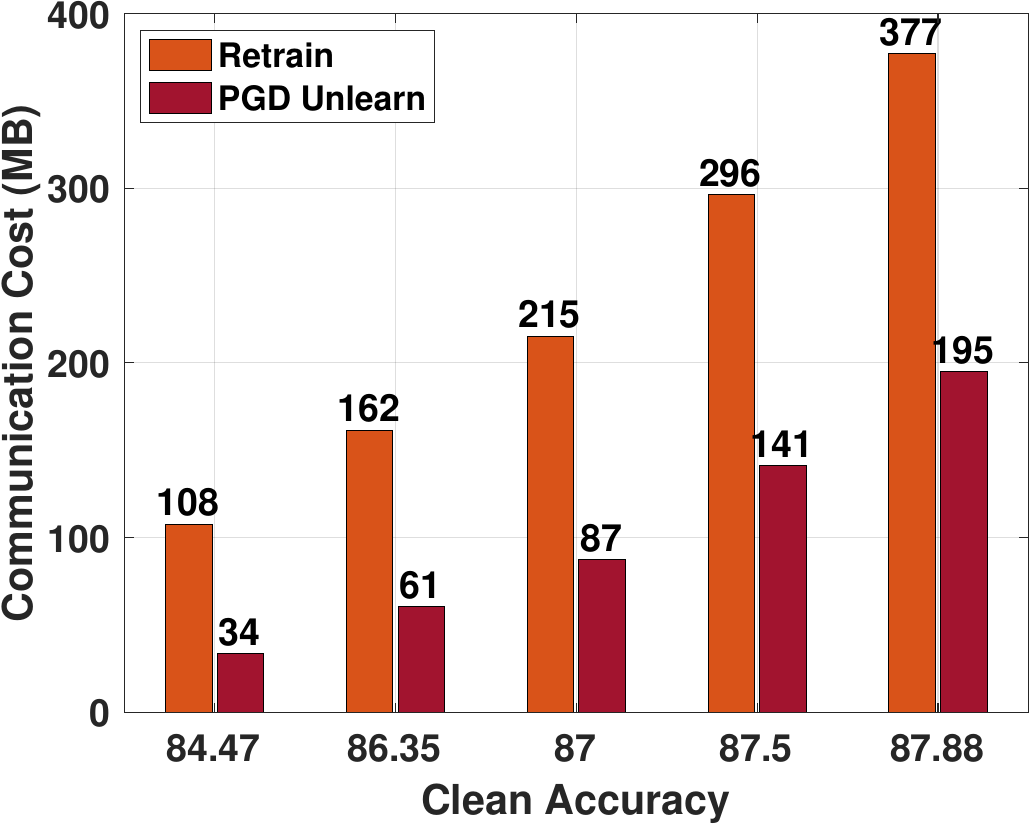}}
    \end{subfigure}\hfill
    \vspace{-5pt}
    \caption{Flipping Scenario: Communication costs (efficiency) of the proposed unlearning method and fully retraining with respect to the clean accuracy (fidelity) in the MNIST and EMNIST dataset for $N=5$ clients.}
    \label{fig:flip_unlearn_comm_costs_num_parties_5}
    \vspace{-5pt}
\end{figure}

\noindent\textbf{Efficiency evaluation:} We compare the communication cost of our approach with retraining to quantify its efficiency. We compute the communication cost of a given method as the total size of the model updates (in MB) that clients send to the server. Figure~\ref{fig:flip_unlearn_comm_costs_num_parties_5} shows the communication cost for various fidelity (clean accuracy) values for $N=5$ clients. In the MNIST dataset, to achieve a clean accuracy of $98.13\%$, our method requires $566$ MB of communication costs, while retraining requires $1039$ MB. Thus, our proposed approach is $1.8 \times$ more efficient than the baseline approach. We obtain similar gains for the EMNIST dataset. Overall, we observe that the proposed unlearning algorithm is up to $5.8 \times$ more efficient than retraining from scratch. In Appendix~\ref{app:flipping_acc_number_rounds}, we provide further evaluation of our method with respect to the number of training rounds. 

Overall, for both unlearning scenarios, we observe that the proposed unlearning method is more efficient in terms of the communication cost on the retained clients than retraining, while achieving comparable fidelity and efficacy. It is worth noting that, even though we do not explicitly measure computation costs, our method also reduces the computation cost as compared to retraining since it requires much fewer rounds than retraining to achieve high performance on the retained data. We believe that lowering the communication and computation burden on retained clients is appealing in practice since these clients are not incentivized to help the target client in unlearning.

\section{Conclusion}\label{sec:conclusion}
We proposed a novel federated unlearning method that can efficiently unlearn the contribution of any client. Our method first performs local unlearning at the client to be erased and, starting with the locally unlearned model, performs a few rounds of FL with the server and remaining clients. Unlike prior federated unlearning works, our method does not require the server (or any other client) to keep track of the history of their parameter updates. We have used the backdoor triggers and flipping to effectively evaluate the performance of the proposed method. We empirically demonstrated the efficacy, fidelity, and efficiency of the proposed unlearning method.

\section*{Acknowledgements}
This work was supported by the European Union's Horizon 2020 research and innovation programme under grant number 951911 – AI4Media.

\bibliography{references}

\begin{thebibliography}{36}
\providecommand{\natexlab}[1]{#1}
\providecommand{\url}[1]{\texttt{#1}}
\expandafter\ifx\csname urlstyle\endcsname\relax
  \providecommand{\doi}[1]{doi: #1}\else
  \providecommand{\doi}{doi: \begingroup \urlstyle{rm}\Url}\fi

\bibitem[Act(2000)]{act2000personal}
Act, P.
\newblock Personal information protection and electronic documents act.
\newblock \emph{Department of Justice, Canada. Full text available at
  http://laws. justice. gc. ca/en/P-8.6/text. html}, 2000.

\bibitem[Baumhauer et~al.(2020)Baumhauer, Sch{\"o}ttle, and
  Zeppelzauer]{baumhauer2020machine}
Baumhauer, T., Sch{\"o}ttle, P., and Zeppelzauer, M.
\newblock Machine unlearning: Linear filtration for logit-based classifiers.
\newblock \emph{arXiv preprint arXiv:2002.02730}, 2020.

\bibitem[Bourtoule et~al.(2021)Bourtoule, Chandrasekaran, Choquette-Choo, Jia,
  Travers, Zhang, Lie, and Papernot]{bourtoule2021machine}
Bourtoule, L., Chandrasekaran, V., Choquette-Choo, C.~A., Jia, H., Travers, A.,
  Zhang, B., Lie, D., and Papernot, N.
\newblock Machine unlearning.
\newblock In \emph{2021 IEEE Symposium on Security and Privacy (SP)}, pp.\
  141--159. IEEE, 2021.

\bibitem[Cao \& Yang(2015)Cao and Yang]{cao2015towards}
Cao, Y. and Yang, J.
\newblock Towards making systems forget with machine unlearning.
\newblock In \emph{2015 IEEE Symposium on Security and Privacy}, pp.\
  463--480. IEEE, 2015.

\bibitem[Carlini et~al.(2021)Carlini, Tram{\`e}r, Wallace, Jagielski,
  Herbert-Voss, Lee, Roberts, Brown, Song, Erlingsson, Oprea, and
  Raffel]{carlini2021extracting}
Carlini, N., Tram{\`e}r, F., Wallace, E., Jagielski, M., Herbert-Voss, A., Lee,
  K., Roberts, A., Brown, T., Song, D., Erlingsson, {\'U}., Oprea, A., and
  Raffel, C.
\newblock Extracting training data from large language models.
\newblock In \emph{30th USENIX Security Symposium (USENIX Security 21)}, pp.\
  2633--2650, August 2021.
\newblock ISBN 978-1-939133-24-3.

\bibitem[Carlini et~al.(2023)Carlini, Ippolito, Jagielski, Lee, Tramer, and
  Zhang]{carlini2023quantifying}
Carlini, N., Ippolito, D., Jagielski, M., Lee, K., Tramer, F., and Zhang, C.
\newblock Quantifying memorization across neural language models.
\newblock In \emph{The Eleventh International Conference on Learning
  Representations}, 2023.

\bibitem[Cohen et~al.(2017)Cohen, Afshar, Tapson, and
  Van~Schaik]{cohen2017emnist}
Cohen, G., Afshar, S., Tapson, J., and Van~Schaik, A.
\newblock Emnist: Extending mnist to handwritten letters.
\newblock In \emph{2017 international joint conference on neural networks
  (IJCNN)}, pp.\  2921--2926. IEEE, 2017.

\bibitem[Du et~al.(2019)Du, Chen, Liu, Oak, and Song]{du2019lifelong}
Du, M., Chen, Z., Liu, C., Oak, R., and Song, D.
\newblock Lifelong anomaly detection through unlearning.
\newblock In \emph{Proceedings of the 2019 ACM SIGSAC Conference on Computer
  and Communications Security}, pp.\  1283--1297, 2019.

\bibitem[Ginart et~al.(2019)Ginart, Guan, Valiant, and Zou]{ginart2019making}
Ginart, A., Guan, M., Valiant, G., and Zou, J.~Y.
\newblock Making ai forget you: Data deletion in machine learning.
\newblock \emph{Advances in Neural Information Processing Systems}, 32, 2019.

\bibitem[Golatkar et~al.(2020{\natexlab{a}})Golatkar, Achille, and
  Soatto]{golatkar2020eternal}
Golatkar, A., Achille, A., and Soatto, S.
\newblock Eternal sunshine of the spotless net: Selective forgetting in deep
  networks.
\newblock In \emph{Proceedings of the IEEE/CVF Conference on Computer Vision
  and Pattern Recognition}, pp.\  9304--9312, 2020{\natexlab{a}}.

\bibitem[Golatkar et~al.(2020{\natexlab{b}})Golatkar, Achille, and
  Soatto]{golatkar2020forgetting}
Golatkar, A., Achille, A., and Soatto, S.
\newblock Forgetting outside the box: Scrubbing deep networks of information
  accessible from input-output observations.
\newblock In \emph{European Conference on Computer Vision}, pp.\  383--398.
  Springer, 2020{\natexlab{b}}.

\bibitem[Graves et~al.(2020)Graves, Nagisetty, and Ganesh]{graves2020amnesiac}
Graves, L., Nagisetty, V., and Ganesh, V.
\newblock Amnesiac machine learning.
\newblock \emph{arXiv preprint arXiv:2010.10981}, 2020.

\bibitem[Gu et~al.(2017)Gu, Dolan-Gavitt, and Garg]{gu2017badnets}
Gu, T., Dolan-Gavitt, B., and Garg, S.
\newblock Badnets: Identifying vulnerabilities in the machine learning model
  supply chain.
\newblock \emph{arXiv preprint arXiv:1708.06733}, 2017.

\bibitem[Guo et~al.(2019)Guo, Goldstein, Hannun, and Van
  Der~Maaten]{guo2019certified}
Guo, C., Goldstein, T., Hannun, A., and Van Der~Maaten, L.
\newblock Certified data removal from machine learning models.
\newblock \emph{arXiv preprint arXiv:1911.03030}, 2019.

\bibitem[Jang et~al.(2023)Jang, Yoon, Yang, Cha, Lee, Logeswaran, and
  Seo]{jang2023knowledge}
Jang, J., Yoon, D., Yang, S., Cha, S., Lee, M., Logeswaran, L., and Seo, M.
\newblock Knowledge unlearning for mitigating privacy risks in language models.
\newblock In \emph{Proceedings of the 61st Annual Meeting of the Association
  for Computational Linguistics (Volume 1: Long Papers)}, pp.\  14389--14408,
  Toronto, Canada, July 2023. Association for Computational Linguistics.
\newblock \doi{10.18653/v1/2023.acl-long.805}.
\newblock URL \url{https://aclanthology.org/2023.acl-long.805}.

\bibitem[Kairouz et~al.(2021)Kairouz, McMahan, Avent, Bellet, Bennis, Bhagoji,
  Bonawitz, Charles, Cormode, Cummings, D’Oliveira, Eichner, Rouayheb, Evans,
  Gardner, Garrett, Gascón, Ghazi, Gibbons, Gruteser, Harchaoui, He, He, Huo,
  Hutchinson, Hsu, Jaggi, Javidi, Joshi, Khodak, Konecný, Korolova,
  Koushanfar, Koyejo, Lepoint, Liu, Mittal, Mohri, Nock, Özgür, Pagh, Qi,
  Ramage, Raskar, Raykova, Song, Song, Stich, Sun, Suresh, Tramèr, Vepakomma,
  Wang, Xiong, Xu, Yang, Yu, Yu, and Zhao]{kairouz2021fl}
Kairouz, P., McMahan, H.~B., Avent, B., Bellet, A., Bennis, M., Bhagoji, A.~N.,
  Bonawitz, K., Charles, Z., Cormode, G., Cummings, R., D’Oliveira, R. G.~L.,
  Eichner, H., Rouayheb, S.~E., Evans, D., Gardner, J., Garrett, Z., Gascón,
  A., Ghazi, B., Gibbons, P.~B., Gruteser, M., Harchaoui, Z., He, C., He, L.,
  Huo, Z., Hutchinson, B., Hsu, J., Jaggi, M., Javidi, T., Joshi, G., Khodak,
  M., Konecný, J., Korolova, A., Koushanfar, F., Koyejo, S., Lepoint, T., Liu,
  Y., Mittal, P., Mohri, M., Nock, R., Özgür, A., Pagh, R., Qi, H., Ramage,
  D., Raskar, R., Raykova, M., Song, D., Song, W., Stich, S.~U., Sun, Z.,
  Suresh, A.~T., Tramèr, F., Vepakomma, P., Wang, J., Xiong, L., Xu, Z., Yang,
  Q., Yu, F.~X., Yu, H., and Zhao, S.
\newblock Advances and open problems in federated learning.
\newblock \emph{Foundations and Trends® in Machine Learning}, 14\penalty0
  (1–2):\penalty0 1--210, 2021.

\bibitem[Krizhevsky et~al.(2009)Krizhevsky, Hinton,
  et~al.]{krizhevsky2009learning}
Krizhevsky, A., Hinton, G., et~al.
\newblock Learning multiple layers of features from tiny images.
\newblock 2009.

\bibitem[Lecun et~al.(1998)Lecun, Bottou, Bengio, and Haffner]{lecun1998mnist}
Lecun, Y., Bottou, L., Bengio, Y., and Haffner, P.
\newblock Gradient-based learning applied to document recognition.
\newblock \emph{Proceedings of the IEEE}, 86\penalty0 (11):\penalty0
  2278--2324, 1998.
\newblock \doi{10.1109/5.726791}.

\bibitem[Lehman et~al.(2021)Lehman, Jain, Pichotta, Goldberg, and
  Wallace]{lehman2021bert}
Lehman, E., Jain, S., Pichotta, K., Goldberg, Y., and Wallace, B.
\newblock Does {BERT} pretrained on clinical notes reveal sensitive data?
\newblock In \emph{Proceedings of the 2021 Conference of the North American
  Chapter of the Association for Computational Linguistics: Human Language
  Technologies}, pp.\  946--959. Association for Computational Linguistics,
  June 2021.

\bibitem[Liu et~al.(2021)Liu, Ma, Yang, Wang, and Liu]{liu2021federaser}
Liu, G., Ma, X., Yang, Y., Wang, C., and Liu, J.
\newblock Federaser: Enabling efficient client-level data removal from
  federated learning models.
\newblock In \emph{2021 IEEE/ACM 29th International Symposium on Quality of
  Service (IWQOS)}, pp.\  1--10. IEEE, 2021.

\bibitem[Liu et~al.(2022)Liu, Xu, Yuan, Wang, and Li]{liu2022the}
Liu, Y., Xu, L., Yuan, X., Wang, C., and Li, B.
\newblock The right to be forgotten in federated learning: An efficient
  realization with rapid retraining.
\newblock In \emph{IEEE INFOCOM 2022 - IEEE Conference on Computer
  Communications}, pp.\  1749--1758, 2022.
\newblock \doi{10.1109/INFOCOM48880.2022.9796721}.

\bibitem[Madry et~al.(2018)Madry, Makelov, Schmidt, Tsipras, and
  Vladu]{madry2018towards}
Madry, A., Makelov, A., Schmidt, L., Tsipras, D., and Vladu, A.
\newblock Towards deep learning models resistant to adversarial attacks.
\newblock In \emph{International Conference on Learning Representations}, 2018.

\bibitem[McMahan et~al.(2017)McMahan, Moore, Ramage, Hampson, and
  y~Arcas]{mcmahan2017communication}
McMahan, B., Moore, E., Ramage, D., Hampson, S., and y~Arcas, B.~A.
\newblock Communication-efficient learning of deep networks from decentralized
  data.
\newblock In \emph{Artificial intelligence and statistics}, pp.\  1273--1282.
  PMLR, 2017.

\bibitem[Neel et~al.(2021)Neel, Roth, and Sharifi-Malvajerdi]{neel2021descent}
Neel, S., Roth, A., and Sharifi-Malvajerdi, S.
\newblock Descent-to-delete: Gradient-based methods for machine unlearning.
\newblock In \emph{Algorithmic Learning Theory}, pp.\  931--962. PMLR, 2021.

\bibitem[Nguyen et~al.(2022)Nguyen, Huynh, Nguyen, Liew, Yin, and
  Nguyen]{nguyen2022survey}
Nguyen, T.~T., Huynh, T.~T., Nguyen, P.~L., Liew, A. W.-C., Yin, H., and
  Nguyen, Q. V.~H.
\newblock A survey of machine unlearning.
\newblock \emph{arXiv preprint arXiv:2209.02299}, 2022.

\bibitem[Nicolae et~al.(2018)Nicolae, Sinn, Tran, Buesser, Rawat, Wistuba,
  Zantedeschi, Baracaldo, Chen, Ludwig, Molloy, and Edwards]{art2018}
Nicolae, M.-I., Sinn, M., Tran, M.~N., Buesser, B., Rawat, A., Wistuba, M.,
  Zantedeschi, V., Baracaldo, N., Chen, B., Ludwig, H., Molloy, I., and
  Edwards, B.
\newblock Adversarial robustness toolbox v1.2.0.
\newblock \emph{CoRR}, 1807.01069, 2018.
\newblock URL \url{https://arxiv.org/pdf/1807.01069}.

\bibitem[Pardau(2018)]{pardau2018california}
Pardau, S.~L.
\newblock The california consumer privacy act: Towards a european-style privacy
  regime in the united states.
\newblock \emph{J. Tech. L. \& Pol'y}, 23:\penalty0 68, 2018.

\bibitem[Sekhari et~al.(2021)Sekhari, Acharya, Kamath, and
  Suresh]{sekhari2021remember}
Sekhari, A., Acharya, J., Kamath, G., and Suresh, A.~T.
\newblock Remember what you want to forget: Algorithms for machine unlearning.
\newblock \emph{Advances in Neural Information Processing Systems}, 34, 2021.

\bibitem[Shokri et~al.(2017)Shokri, Stronati, Song, and
  Shmatikov]{shokri2017membership}
Shokri, R., Stronati, M., Song, C., and Shmatikov, V.
\newblock Membership inference attacks against machine learning models.
\newblock In \emph{2017 IEEE symposium on security and privacy (SP)}, pp.\
  3--18. IEEE, 2017.

\bibitem[Thudi et~al.(2021)Thudi, Deza, Chandrasekaran, and
  Papernot]{thudi2021unrolling}
Thudi, A., Deza, G., Chandrasekaran, V., and Papernot, N.
\newblock Unrolling sgd: Understanding factors influencing machine unlearning.
\newblock \emph{arXiv preprint arXiv:2109.13398}, 2021.

\bibitem[Voigt \& Von~dem Bussche(2017)Voigt and Von~dem Bussche]{voigt2017eu}
Voigt, P. and Von~dem Bussche, A.
\newblock The eu general data protection regulation (gdpr).
\newblock \emph{A Practical Guide, 1st Ed., Cham: Springer International
  Publishing}, 10\penalty0 (3152676):\penalty0 10--5555, 2017.

\bibitem[Wang et~al.(2022)Wang, Guo, Xie, and Qi]{wang2022federated}
Wang, J., Guo, S., Xie, X., and Qi, H.
\newblock Federated unlearning via class-discriminative pruning.
\newblock In \emph{Proceedings of the ACM Web Conference 2022}, pp.\  622--632,
  2022.

\bibitem[Warnecke et~al.(2023)Warnecke, Pirch, Wressnegger, and
  Rieck]{warnecke2023machine}
Warnecke, A., Pirch, L., Wressnegger, C., and Rieck, K.
\newblock Machine unlearning of features and labels.
\newblock In \emph{30th Annual Network and Distributed System Security
  Symposium, {NDSS} 2023, San Diego, California, USA, February 27 - March 3,
  2023}. The Internet Society, 2023.

\bibitem[Wu et~al.(2022)Wu, Zhu, and Mitra]{wu2022federated}
Wu, C., Zhu, S., and Mitra, P.
\newblock Federated unlearning with knowledge distillation.
\newblock \emph{arXiv preprint arXiv:2201.09441}, 2022.

\bibitem[Xu et~al.(2023)Xu, Zhu, Zhang, Zhou, and Yu]{xu2023machine}
Xu, H., Zhu, T., Zhang, L., Zhou, W., and Yu, P.~S.
\newblock Machine unlearning: A survey.
\newblock \emph{ACM Computing Surveys}, 56\penalty0 (1):\penalty0 1--36, 2023.

\bibitem[Yeom et~al.(2018)Yeom, Giacomelli, Fredrikson, and
  Jha]{yeom2018privacy}
Yeom, S., Giacomelli, I., Fredrikson, M., and Jha, S.
\newblock Privacy risk in machine learning: Analyzing the connection to
  overfitting.
\newblock In \emph{2018 IEEE 31st computer security foundations symposium
  (CSF)}, pp.\  268--282. IEEE, 2018.

\end{thebibliography}
\bibliographystyle{icml2022}

\appendix
\onecolumn
\section{Details on Hyperparameters}\label{app:hyperparameters}

During FL training, we use the SGD optimizer with the following hyperparameters:
\begin{itemize}[noitemsep]
    \item Momentum $\beta = 0.9$
    \item Learning rate $\eta = 0.01$
    \item Batch size $B = 128$
    \item Number of epochs $E = 1$
    \item Aggregation algorithm: FedAvg (described in Section 3)
\end{itemize} 

For PGD-based unlearning, we use the SGD optimizer with the following hyperparameters:
\begin{itemize}[noitemsep]
    \item Momentum $\beta = 0.9$
    \item Learning rate $\eta_u = 0.01$
    \item Batch size $B_u = 1024$
    \item Number of epochs $E_u = 5$
    \item $\ell_2$-norm ball radius $\delta$ is set to be one third of the average Euclidean distance between $\vect{w}_{\textrm{ref}}$ and a random model, where the average is computed over $10$ random models. This value is selected because we want the model to stay closer to the reference model than a random model.
    \item Early stopping threshold $\tau$: Computed via a grid search over the interval $[2,6]$
    \item Gradient $\ell_2$-clipping is employed with radius $5$
\end{itemize}

For FL post-training after unlearning, we use the SGD optimizer with the following hyperparameters:
\begin{itemize}[noitemsep]
    \item Momentum $\beta = 0.9$
    \item Learning rate $\eta_p = 0.01$
    \item Batch size $B_p = 128$
    \item FL rounds $T_p:$ see Figure 5 in Section 5.1, Appendices~\ref{app:backdoor_acc_number_rounds}, and~\ref{app:flipping_acc_number_rounds}
    \item Aggregation algorithm: FedAvg 
\end{itemize}

\subsection*{Schematic overview of local unlearning} 
\begin{figure}[ht!]
    \begin{center}
    \includegraphics[scale=0.7]{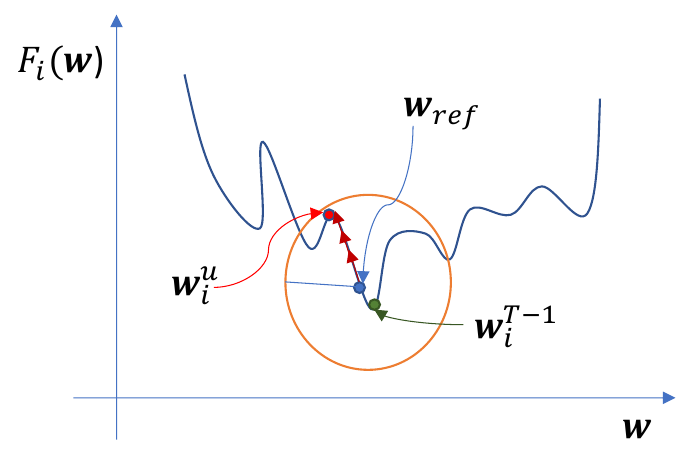}
    \caption{A schematic to illustrate the main idea of the local unlearning phase in Algorithm~1.}
    \label{fig:PGA-schematic}
    \end{center}
\end{figure}

Local unlearning at the client is formulated as a constrained optimization problem where given a trained model and the data to be removed, the algorithm obtains an updated model in the vicinity of the original model which exhibits poor performance on the deleted data. The neighborhood is modelled as an $\ell_2$-norm ball of radius $\delta$ around $\vect{w}_{\textrm{ref}}$ as shown in Figure~\ref{fig:PGA-schematic}. In an FL setting, a natural choice of such a $\vect{w}_{\textrm{ref}}$ can be obtained by removing the appropriately scaled last update from the current global model. The algorithm proceeds by taking gradient steps in the direction that maximizes the empirical loss on its local data using Projected Gradient Descent (PGD), which is represented by the red arrows in Figure~\ref{fig:PGA-schematic}. By constraining the parameters to remain within the feasible region of this optimization problem, the algorithm ensures the retention of performance with respect to other clients. The choice of $\ell_2$-norm ball for this feasible set is also consistent with the observations in~\cite{thudi2021unrolling} which argue the use of Euclidean distance on model parameters for unlearning verification. The combined choice of $\delta$ and $\vect{w}_{\textrm{ref}}$ can be used to balance the performance trade-off between deleted and retained clients. As we noted in our experiments, even with a slight compromise in clean accuracy (retention) $\vect{w}^{u}_i$ quickly recovers after a few steps of FL  post-training (second phase in Algorithm~1).

\section{Membership Inference Attacks}\label{app:mem_inf}
In this section, we briefly describe Shokri's attack and Yeom's attack.

\noindent\paragraph{Shokri's attack} \citet{shokri2017membership} were the first to propose a membership inference attack on machine learning (ML) models. Let $M$ be the target model trained on dataset $D$. The main intuition of the attack is that ML models tend to behave differently on the training data compared to the data that they have not seen. It is assumed that the attacker knows the type and the architecture of the model $M$ and has access to some data $D_S$ that comes from the same underlying distribution as training data $D$. Thus, the attacker can train multiple shadow models $M_{S_i}$ (one per class) that mimic the behavior of the target model. For the shadow models, the attacker has their training and test datasets and thus knows the groundtruth of the membership of the training and test data samples. Based on this, the attacker trains multiple attack models $M_{A_i}$ (one per class) by using as input the posteriors returned by the corresponding shadow model and as labels their membership. Finally, when the attacker wants to determine the membership of a target data sample, they query the target model $M$, obtain its posterior probability, and with that query the corresponding attack model $M_{A_i}$ to obtain the membership prediction.

\noindent\paragraph{Yeom's attack} \citet{yeom2018privacy} assume that the attacker has white-box access to the target model $M$ and knows its average training loss. To determine the membership of a target data sample, the attacker computes the loss of the model $M$ on the input data sample and compares this value to the average training loss of the model $M$. If the loss of the target data sample is smaller than the average training loss, then it is classified as a member, otherwise as a non-member.

\section{Efficiency Evaluation in Backdoors}\label{app:backdoor_acc_number_rounds}
\begin{figure*}[h!]
    \centering
    \begin{subfigure}[MNIST]{\includegraphics[scale=0.28]{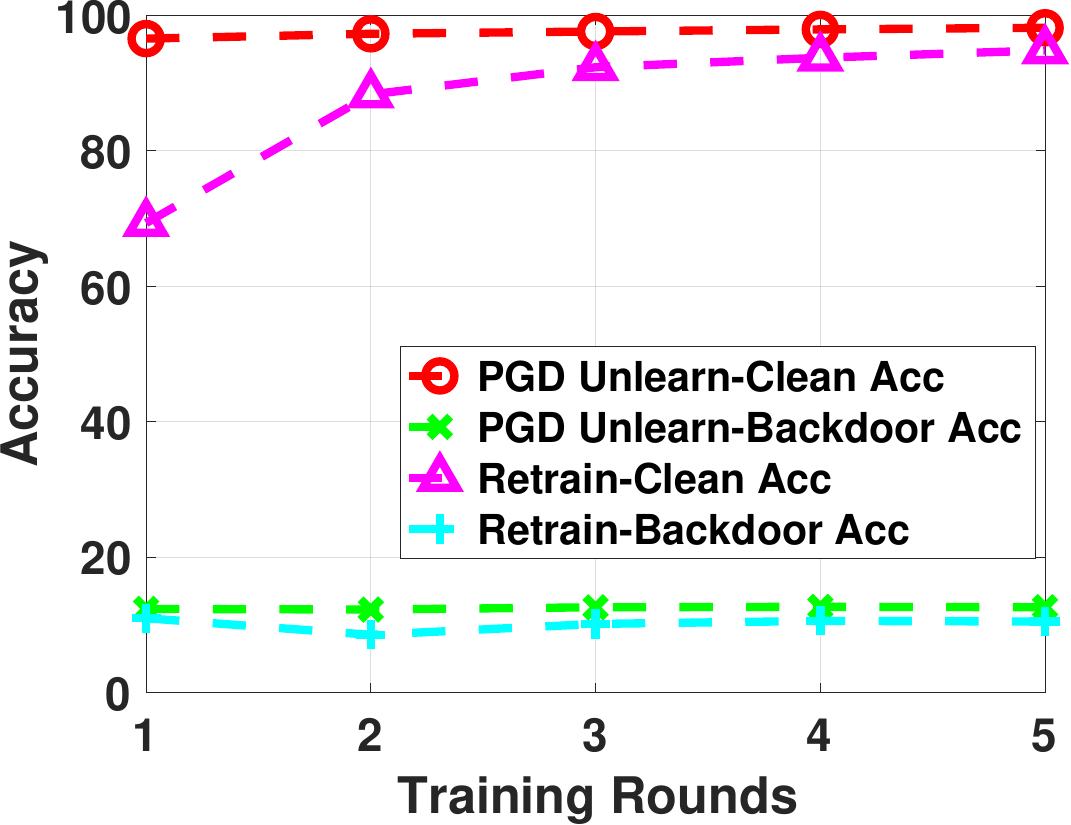}}
    \end{subfigure}\hfill
    \begin{subfigure}[EMNIST]{\includegraphics[scale=0.28]{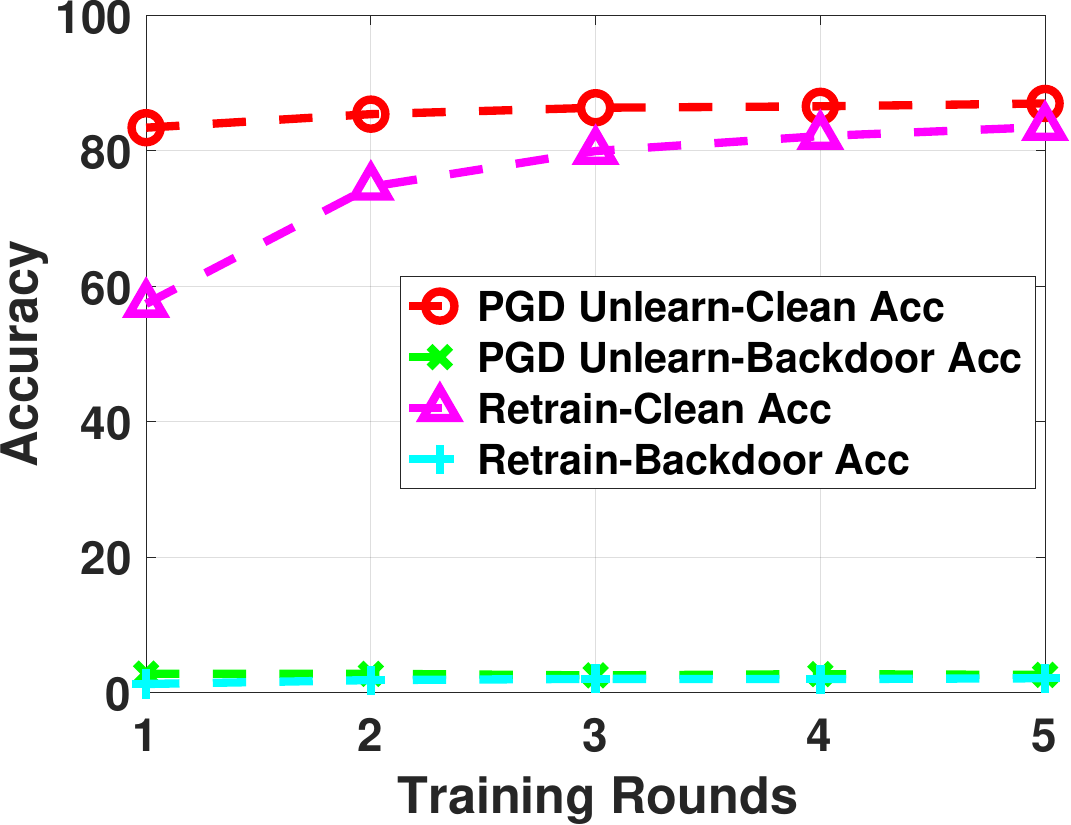}}
    \end{subfigure}\hfill
    \begin{subfigure}[CIFAR-10]{\includegraphics[scale=0.28]{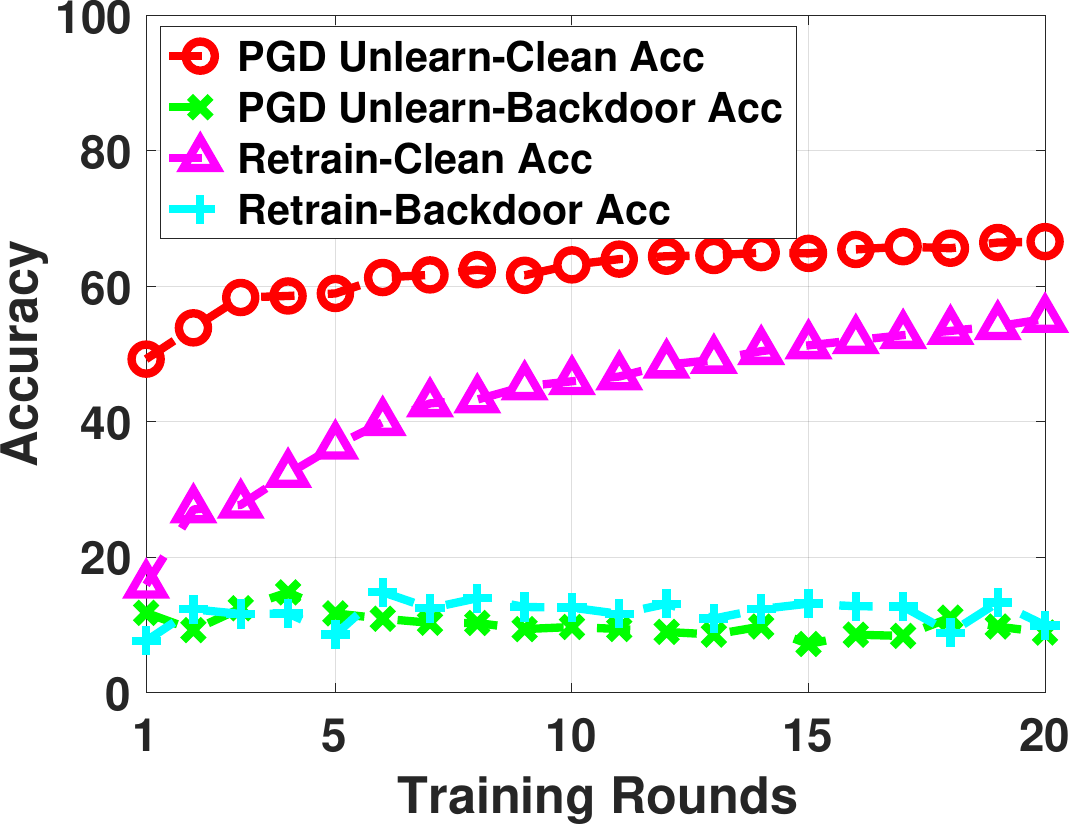}}
    \end{subfigure}\hfill
    \caption{Backdoor Scenario: Clean accuracy (fidelity) and backdoor accuracy (efficacy) of the proposed unlearning method and full FL retraining from scratch with respect to the number of rounds in each dataset for $N=10$ clients.}
    \label{fig:unlearn_client_rounds_num_parties_10}
\end{figure*}

In Figure 5 (in Section~\ref{sec:backdoored_images}), we showed the comparison of the clean accuracy (fidelity) and backdoor accuracy (efficacy) of the proposed PGD-based unlearning method to the gold standard (retraining from scratch) with respect to the number of FL training rounds for $N=5$ clients. Here, in Figure~\ref{fig:unlearn_client_rounds_num_parties_10}, we show this comparison for $N=10$ clients. Note that, for our method, the FL training starts with the locally unlearned model that the target client has obtained using the projected gradient descent (as discussed in Section~\ref{sec:unlearning_method}), while for the baseline of retraining, the FL training starts with a randomly initialized model. We observe that local unlearning at the target client bootstraps the unlearning process, and requires substantially fewer number of FL rounds than retraining from scratch. For instance, after one round of post-training, the proposed method reaches a clean accuracy of $96.6\%$ and a backdoor accuracy of $12.35\%$ in the MNIST dataset. The baseline method requires more than $5$ FL rounds to achieve similar performance, showing the efficiency of the proposed unlearning method. 

\section{Efficiency Evaluation in Flipping}\label{app:flipping_acc_number_rounds}
\begin{figure}[ht!]
    \centering
    \hfill
    \begin{subfigure}[MNIST]{\includegraphics[scale=0.33]{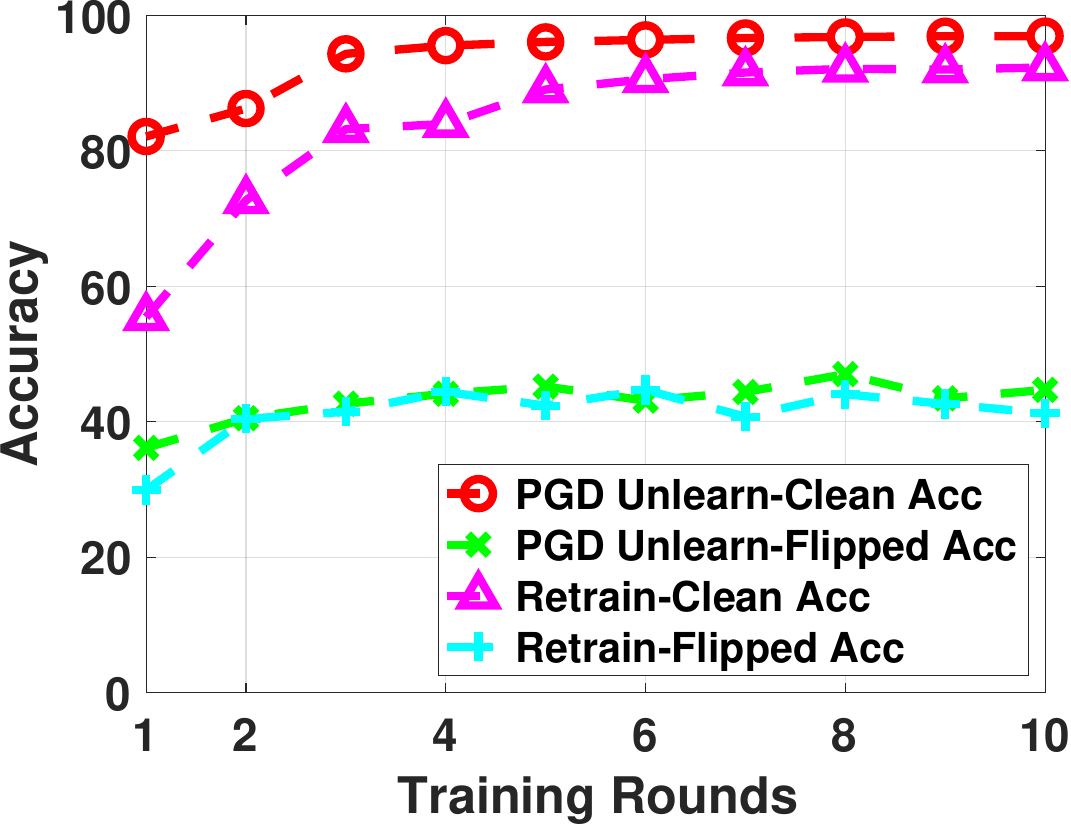}}
    \end{subfigure}\hfill
    \begin{subfigure}[EMNIST]{\includegraphics[scale=0.33]{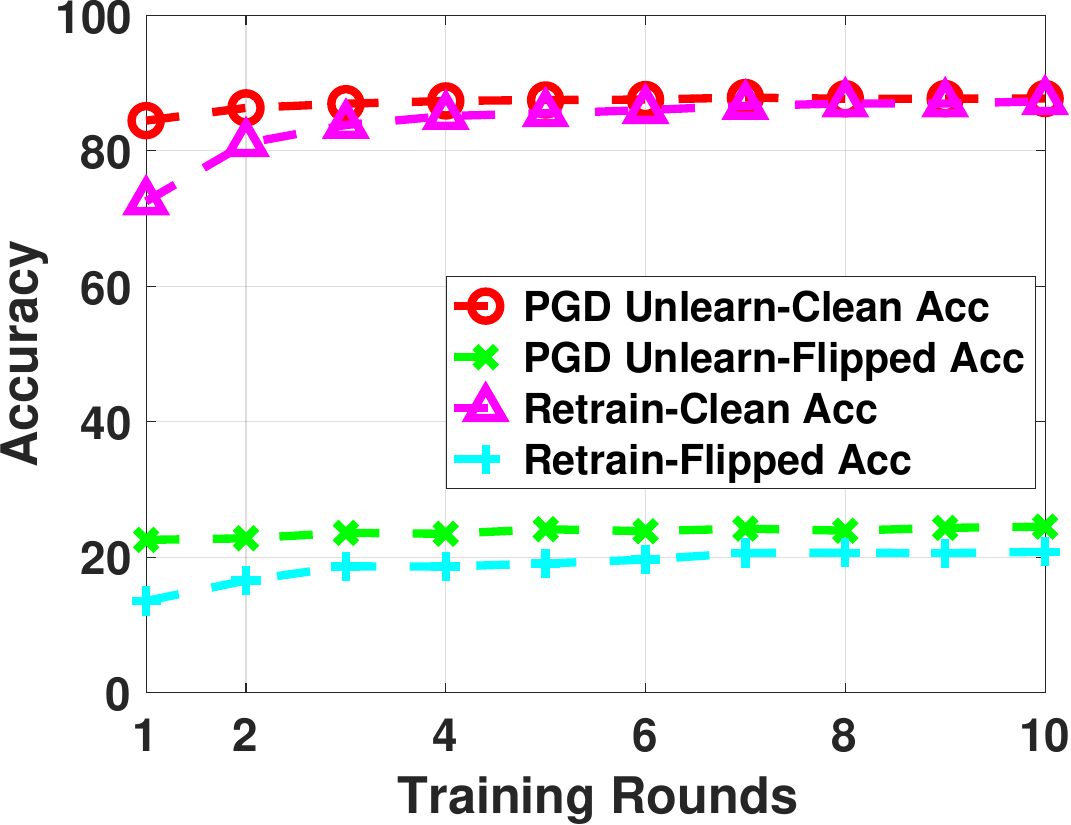}} \hfill
    \end{subfigure}\hfill
    % \vspace{-8pt}
    \caption{Flipping Scenario: Clean accuracy (fidelity) and flipped accuracy (efficacy) of the proposed unlearning method and full FL retraining from scratch with respect to the number of rounds in the MNIST and EMNIST datasets for $N=5$ clients.}
    \label{fig:unlearn_flipping_num_parties_5}
    % \vspace{-15pt}
\end{figure}

\begin{figure}[h!]
    \centering
    \hfill
    \begin{subfigure}[MNIST]{\includegraphics[scale=0.33]{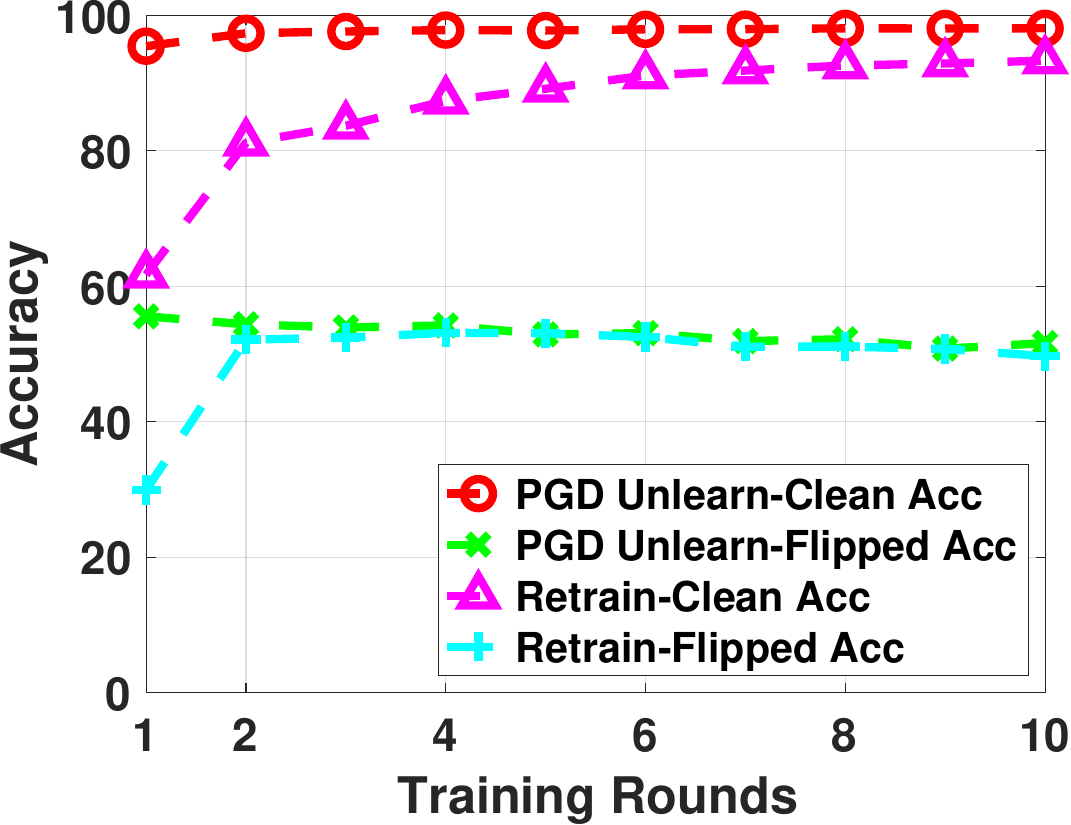}}
    \end{subfigure}\hfill
    \begin{subfigure}[EMNIST]{\includegraphics[scale=0.33]{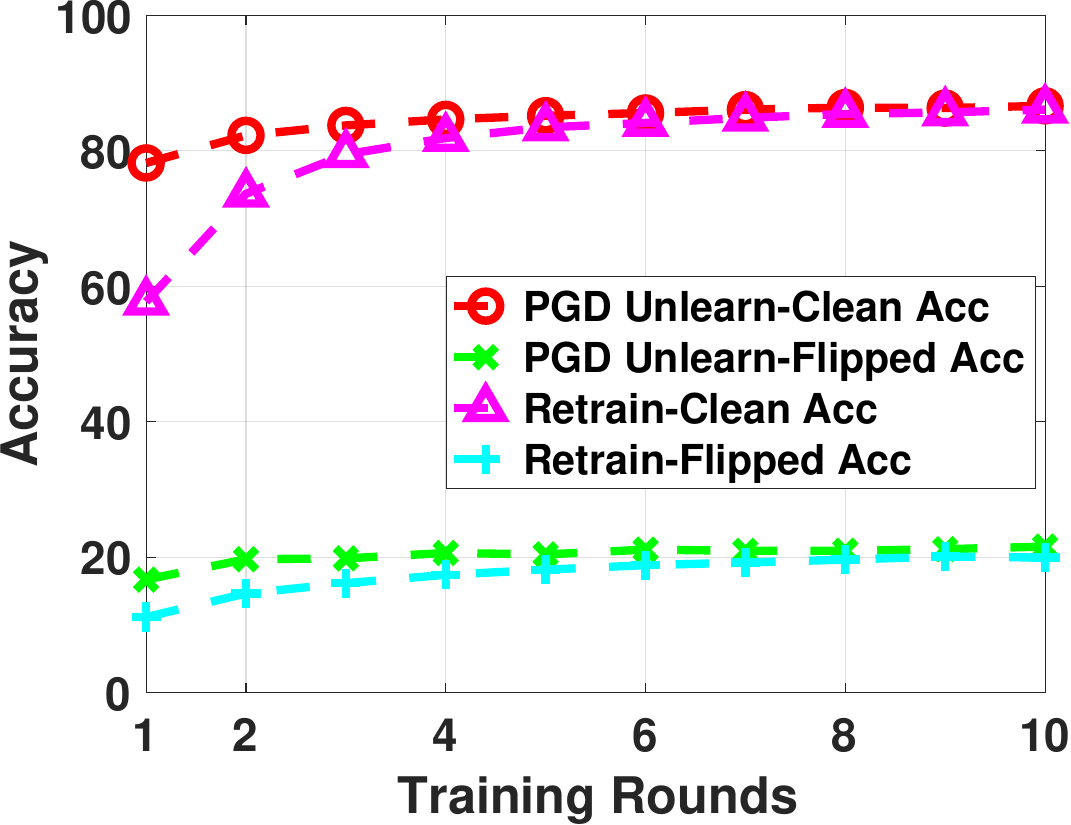}} \hfill
    \end{subfigure}\hfill
    \caption{Flipping Scenario: Clean accuracy (fidelity) and flipped accuracy (efficacy) of the proposed unlearning method and full FL retraining from scratch with respect to the number of rounds in the MNIST and EMNIST datasets for $N=10$ clients.}
    \label{fig:unlearn_flipping_num_parties_10}
\end{figure}

In Figures~\ref{fig:unlearn_flipping_num_parties_5} and~\ref{fig:unlearn_flipping_num_parties_10}, we show the clean accuracy (fidelity) and flipped accuracy (efficacy) of our method and the fully FL retraining (gold standard) with respect to the number of FL training rounds (FL post-training rounds for the proposed method) for $N=5$ and $N=10$ clients, respectively. We observe that local unlearning at the target client bootstraps the unlearning process, and requires substantially fewer number of FL rounds than retraining from scratch. For example, for $N=10$ in the MNIST dataset, the proposed  unlearning method achieves a clean accuracy of $95.47\%$ and a flipped accuracy of $55.57\%$ after one round of post-training, while the baseline method requires more than $10$ FL rounds to achieve similar performance. These results show the efficiency of the proposed unlearning method. 

\end{document}